%% file: main.tex
\documentclass[runningheads]{llncs}

\usepackage{eccv}
\usepackage{multirow} 
\usepackage{arydshln}
\usepackage[table]{xcolor}
\usepackage{pifont}
\usepackage{caption}
\usepackage{makecell}
\usepackage{eccvabbrv}

\usepackage{graphicx}
\usepackage{booktabs}

\usepackage[accsupp]{axessibility}  % Improves PDF readability for those with disabilities.

\usepackage[pagebackref,breaklinks,colorlinks,citecolor=eccvblue]{hyperref}
\usepackage{orcidlink}
\usepackage{amsmath,amssymb}
\usepackage{subcaption}
\usepackage{algorithm}
\usepackage{algorithmic}
\usepackage{enumitem}
\usepackage{bm}

\newcommand{\gain}[2]{\shortstack{#1\\\textcolor{red}{\scriptsize(#2)}}}

\begin{document}

\title{Revisiting Weakly-Supervised Video Scene Graph Generation via Pair Affinity Learning} 
\titlerunning{WS-VSGG via Pair Affinity Learning}
%\titlerunning{WS-VSGG via Pair Affinity Learning}
\author{Minseok Kang\inst{1} \and
Minhyeok Lee\inst{1} \and
Minjung Kim\inst{2} \and
Jungho Lee\inst{1} \and\\
Donghyeong Kim\inst{1} \and
Sungmin Woo\inst{1} \and
Inseok Jeon\inst{1} \and
Sangyoun Lee\inst{1}}
\authorrunning{M.~Kang et al.}

\institute{Yonsei University \and LG Electronics \\ 
\email{louis0503@yonsei.ac.kr}}

\maketitle
\vspace{-0.5cm}

\begin{abstract}
Weakly-supervised video scene graph generation (WS-VSGG) aims to parse video content into structured relational triplets without bounding box annotations and with only sparse temporal labeling, significantly reducing annotation costs. Without ground-truth bounding boxes, these methods rely on off-the-shelf detectors to generate object proposals, yet largely overlook a fundamental discrepancy from fully-supervised pipelines. Fully-supervised detectors implicitly filter out non-interactive objects, while off-the-shelf detectors indiscriminately detect all visible objects, overwhelming relation models with noisy pairs. We address this by introducing a learnable pair affinity that estimates the likelihood of interaction between subject–object pairs. Through Pair Affinity Learning and Scoring (PALS), pair affinity is incorporated into inference-time ranking and further integrated into contextual  reasoning through Pair Affinity Modulation (PAM), enabling the model  to suppress non-interactive pairs and focus on relationally meaningful ones. To provide cleaner supervision for pair affinity learning, we further propose Relation-Aware Matching (RAM), which leverages vision-language grounding to resolve class-level ambiguity in pseudo-label generation. Extensive experiments on Action Genome demonstrate that our approach consistently yields substantial improvements across different baselines and backbones, achieving state-of-the-art WS-VSGG performance.
\end{abstract}

\section{Introduction}
\label{sec:intro}

Video scene graph generation (VSGG) aims to parse video content into structured triplets in the form of $\langle$subject, predicate, object$\rangle$, capturing spatial and temporal interactions between entities and supporting downstream tasks such as video understanding~\cite{ji2020action, rodin2024action}, visual reasoning~\cite{lu2016visual, hudson2019gqa, johnson2018image}, and video question answering~\cite{xiao2022video, grunde2021agqa, wang2024language}.
Despite its significance, fully-supervised VSGG (FS-VSGG) methods face a critical scalability bottleneck: every frame requires exhaustive bounding box annotations for all objects together with their relationship triplets. 

\begin{figure}[t]
    \centering
    \includegraphics[width=1.0\textwidth]{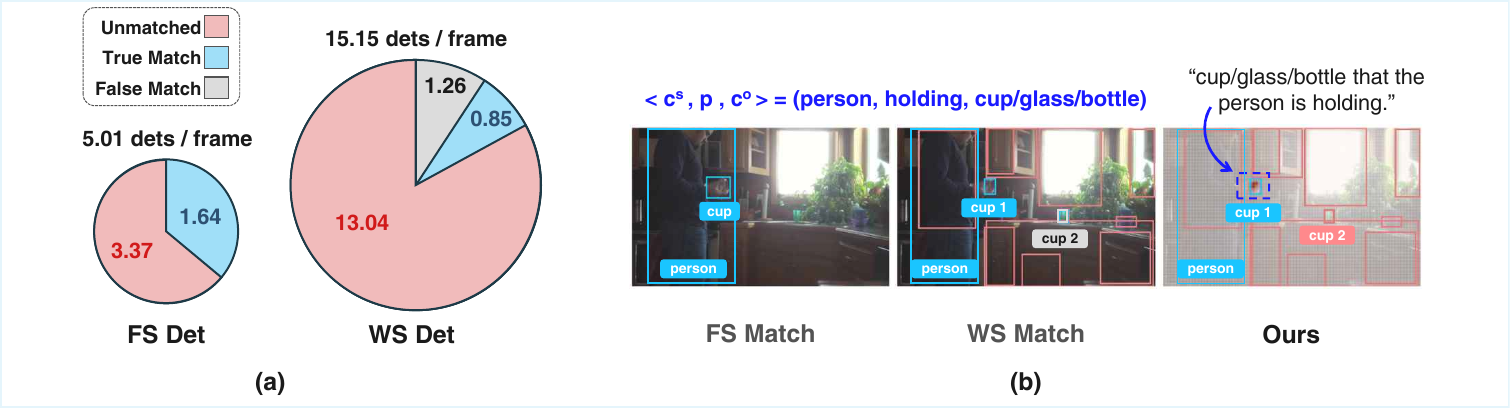}
    \vspace{-0.5cm}
    \caption{\textbf{Detection distribution and matching in FS-VSGG and WS-VSGG.} \textbf{(a)} Per-frame detection statistics in the weakly annotated training set. 
    Detections are categorized as True Match (spatial overlap with a ground-truth), False Match (class-matched without overlap), or Unmatched. \textbf{(b)} Detection and matching comparison for the triplet $\langle$person, holding, cup/glass/bottle$\rangle$.  Fully-supervised detector produces relation-relevant proposals, whereas off-the-shelf detector generates many irrelevant objects, leading to false matches  (e.g., cup 2) under class-level matching. Our method retains only the interaction-consistent instance (cup 1).}
    \vspace{-0.4cm}
    \label{fig:teaser}
\end{figure}

To alleviate this burden, recent works have explored   weakly-supervised VSGG (WS-VSGG), where supervision is provided in a reduced or indirect form.  Representative approaches include PLA~\cite{chen2022video}, which leverages sparse temporal supervision with unlocalized triplets annotated only on the middle frame, where each triplet specifies the subject class, predicate, and object class without bounding boxes. Without ground-truth bounding boxes, WS-VSGG methods inevitably rely on off-the-shelf object detector~\cite{zhang2021vinvl} to generate object proposals, leading most approaches to adopt two-stage architectures~\cite{cong2021spatial, feng2023exploiting} that first detect objects and then predict relations. However, existing WS-VSGG methods~\cite{chen2022video, xu2025trkt} largely adopt this pipeline without accounting for a fundamental shift in detection characteristics between the fully-supervised and weakly-supervised regimes.
As illustrated in Figure~\ref{fig:teaser}(a), a detector fine-tuned on scene graph datasets produces a curated set of proposals centered around relationally relevant objects. An off-the-shelf detector in WS-VSGG, by contrast, produces a far larger set of proposals dominated by objects irrelevant to any interaction. This shifts the burden of filtering non-interactive pairs from the detector to the relation prediction model. 

In conventional two-stage WS-VSGG training pipelines, detector outputs are matched to triplet annotations to assign relation labels. Given unlocalized triplets $\mathcal{G}^u$ for the middle frame $t$ and detected object proposals $\mathcal{D}_t$, detections are matched to triplets based on class identity to construct pseudo-localized training pairs. As illustrated in Fig.~2(a), only matched pairs $\mathcal{P}^+$ receive supervision while unmatched pairs $\mathcal{P}^-$ are entirely discarded. While this convention is viable in FS-VSGG, it creates a severe train-test distribution gap in WS-VSGG. The relation prediction model is trained exclusively on interactive pairs, yet at inference it must operate over a detection space dominated by non-interactive pairs that were never observed during training.

\begin{figure}[t]
    \centering
    \includegraphics[width=1.0\textwidth]{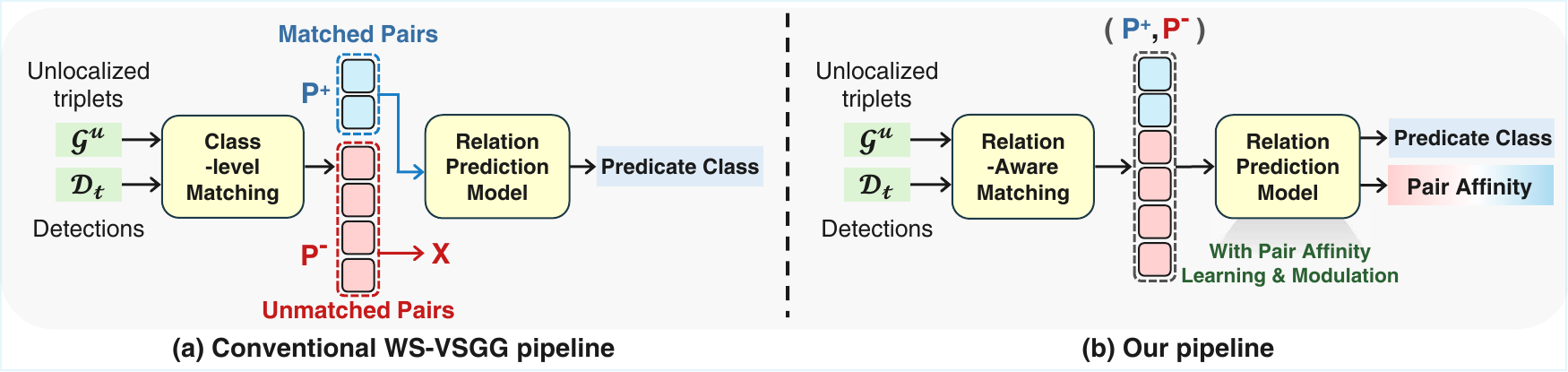}
    \vspace{-0.3cm}
    \caption{\textbf{Comparison of WS-VSGG training pipelines.} $\mathcal{G}^u$: annotation of unlocalized triplets, $\mathcal{D}_t$: detected object proposals. (a) Only matched pairs $\mathcal{P}^+$ are used for training; unmatched pairs $\mathcal{P}^-$ are discarded. (b) RAM refines the matching, and both $\mathcal{P}^+$ and $\mathcal{P}^-$ are used to learn predicate classification and pair affinity jointly.}
    \label{fig:pipeline}
    \vspace{-0.7cm}
\end{figure}

We address this by introducing a learnable \textbf{pair affinity} that captures the boundary between interactive and non-interactive pairs, whose scalar output measures whether a subject–object pair actively participates in an interaction regardless of the specific predicate. To jointly learn pair
affinity alongside predicate classification, we introduce separate
representations for the two objectives within the relation prediction
model, each with distinct supervision signals. Unlike conventional
training pipelines that discard unmatched pairs entirely
(Figure~\ref{fig:pipeline}(a)), we retain them as negative supervision
for pair affinity learning, together with matched pairs as positive
(Figure~\ref{fig:pipeline}(b)). At inference, the learned pair affinity score is incorporated into
triplet ranking to suppress non-interactive pairs.

While pair affinity score suppresses non-interactive pairs at the output level, the model's internal contextual reasoning remains susceptible to their dominance. Modern VSGG architectures~\cite{cong2021spatial, feng2023exploiting, khandelwal2024flocode, nag2023unbiased} perform contextual reasoning through spatial and temporal attention layers, where each pair refines its representation by attending to other pairs, yet non-interactive pairs equally participate and dilute relational context. To prevent non-interactive pairs from dominating this process, we propose \textbf{Pair Affinity Modulation (PAM)}, which gates attention between pairs based on their affinity, allowing interactive pairs to preferentially attend to each other. PAM is applied to both spatial and temporal attention layers, improving contextual reasoning within and across frames as a plug-and-play module compatible with existing attention-based architectures.

Both pair affinity learning and modulation are supervised using the matched\-/unmatched partition. 
Since only unlocalized triplets are available, matching relies solely on class identity, which can generate incorrect pseudo labels when multiple instances of the same class coexist in a frame (Fig.~1(b)). 
These false matches corrupt both predicate classification and pair affinity supervision by introducing spurious positive labels for non-interacting pairs.

We propose \textbf{Relation-Aware Matching (RAM)} to resolve this class-level matching ambiguity by leveraging vision-language (VL) models~\cite{li2022grounded,liu2024grounding} to ground relational descriptions to specific object instances. 
RAM constructs relation-aware text queries from each triplet and uses cross-modal attention from the VL model to identify the most relation-consistent detection. As illustrated in Figure~\ref{fig:teaser}(b), class-level matching indiscriminately assigns all same-class detections as positive, including non-interactive instances (cup 2), while RAM retains only the relation-consistent one (cup 1). RAM is a one-time preprocessing step that refines pseudo-labels for training, providing cleaner supervision for pair affinity learning with no overhead at inference.

Together, our contributions form a unified framework that addresses the detection distribution shift in WS-VSGG. Extensive experiments on the Action Genome dataset demonstrate consistent improvements across multiple baselines and backbones, achieving state-of-the-art performance.

\section{Related Work}
\label{sec:related}

\subsection{Fully-Supervised Video Scene Graph Generation}
VSGG extends image-level scene graph generation to the temporal domain,
producing scene graphs that capture dynamic relationships between objects
across video frames. Existing methods operate at different granularities:
video-level methods~\cite{shang2017video, tsai2019video, shang2021video}
generate a global scene graph for an entire clip, while frame-level methods
have been actively explored since the introduction of Action Genome
(AG)~\cite{ji2020action}, which provides dense frame-level scene graph
annotations over Charades videos and has served as the predominant
benchmark for this setting. Among frame-level
methods~\cite{teng2021target, nag2023unbiased, khandelwal2024flocode,
wang2024oed}, STTran~\cite{cong2021spatial} captures within-frame object
interactions via a spatial transformer and propagates relational context
across adjacent frames via a temporal transformer.
TRACE~\cite{teng2021target} adaptively aggregates context based on target
relevance, and DSG-DETR~\cite{feng2023exploiting} captures long-term
temporal context through class-wise inter-frame graphs. More recently, OED~\cite{wang2024oed} reformulates the task as a set prediction problem, enabling one-stage end-to-end generation. Beyond spatio-temporal modeling, TEMPURA~\cite{nag2023unbiased} and FloCoDe~\cite{khandelwal2024flocode} address the long-tailed predicate distribution through memory-guided training and correlation debiasing, respectively. STTran and DSG-DETR serve as the backbone relation prediction models in our experiments.

\subsection{Weakly-Supervised Video Scene Graph Generation}
To alleviate the heavy annotation burden of fully-supervised methods,
weakly-supervised approaches have been proposed.
PLA~\cite{chen2022video} formulates a weakly-supervised setting by
annotating only the middle frame of each video clip with unlocalized
triplets. Since bounding boxes are unavailable, an off-the-shelf
detector generates object proposals for all frames, which are then matched
to the triplet annotations via class identity to produce pseudo-localized
labels. Although annotations are provided only for the middle frame, PLA
propagates supervision to neighboring frames by treating detections in
those frames that share the same class as the annotated entities as
additional training samples, enabling frame-level scene graph generation
across the entire clip. Building upon this pipeline,
TRKT~\cite{xu2025trkt} observes that off-the-shelf detectors trained on
static images produce suboptimal proposals in dynamic scenes and refines
the detector with temporal information. More recently,
NL-VSGG~\cite{kim2025weakly} further reduces annotation costs by
leveraging video captions instead of explicit triplet annotations.
Notably, these methods commonly adopt the above two-stage
architectures~\cite{cong2021spatial, feng2023exploiting} without adapting
the relation prediction model to the weakly-supervised regime. Our work
fills this gap by equipping the relation prediction model with pair
affinity-aware mechanisms tailored for the weakly-supervised setting.

\section{Method}
\label{sec:method}

\subsection{Problem Formulation}

In FS-VSGG, a model is trained to generate a scene graph $\mathcal{G}_t = \{(s_i, p_i, o_i)\}_{i=1}^{N_t}$ for each frame $t$ in a video clip, where $s_i$, $p_i$, and $o_i$ denote the subject, predicate, and object of the $i$-th triplet, respectively. Each subject and object is associated with a bounding box and a class label, and $N_t$ is the number of annotated triplets in frame $t$. Training requires localized scene graph annotations on every frame, consisting of bounding boxes, object categories, and pairwise relation labels.

In the weakly-supervised setting we consider, following PLA~\cite{chen2022video}, only unlocalized triplets $\mathcal{G}^u = \{(c_i^s, p_i, c_i^o)\}_{i=1}^{M}$ are provided for the middle frame of each video clip, where $c_i^s$ and $c_i^o$ are the subject and object class labels without bounding boxes, $p_i$ is the predicate, and $M$ is the number of annotated triplets. An off-the-shelf object detector~\cite{zhang2021vinvl} is employed to generate object proposals $\mathcal{D}_t = \{(b_j, c_j, \text{conf}_j)\}_{j=1}^{K_t}$ for each frame $t$, where $b_j$, $c_j$, and $\text{conf}_j$ denote the bounding box, predicted class, and confidence score of the $j$-th detection, respectively. These detections are then matched with the unlocalized triplets to produce pseudo-localized scene graphs for training.

\subsection{Overview}
As discussed in Section~\ref{sec:intro}, directly transferring a
fully-supervised two-stage pipeline to the weakly-supervised setting leaves the relation prediction model unable to distinguish interactive pairs from non-interactive ones. Our method addresses this by equipping the model with a learnable pair affinity, through a unified pipeline consisting of three components. Since pair affinity learning is supervised by the matched/unmatched partition, we first describe \textbf{Relation-Aware Matching (RAM)} (Section~\ref{sec:ram}), a preprocessing step that produces this partition by leveraging cross-modal grounding to select the most relation-consistent object among same-class candidates.
\textbf{Pair Affinity Learning and Scoring (PALS)}
(Section~\ref{sec:pals}) then trains a binary interaction score on both
matched and unmatched pairs and incorporates it into inference-time triplet
ranking. Finally, \textbf{Pair Affinity Modulation (PAM)}
(Section~\ref{sec:pam}) uses pair affinity to modulate spatial and temporal
attention, enabling effective contextual reasoning even when non-interactive
pairs dominate the input.

\subsection{Relation-Aware Matching}
\label{sec:ram}

In WS-VSGG, where bounding box annotations are unavailable, pseudo-localized supervision must be constructed by matching unlocalized triplet annotations with detected objects. Since each unlocalized triplet $(c^s, p, c^o)$ provides only class-level identity without spatial information, existing methods assign all detections whose predicted class matches a given entity as positive pseudo labels. Formally, for each entity $c \in \{c^s, c^o\}$, the set of class-matched
detections is $\mathcal{D}_{c} = \{d_j \in \mathcal{D}_t : c_j = c\}$, which inevitably includes false matches when multiple instances of the same class coexist in a frame (e.g., cup 2 in Figure~\ref{fig:teaser}(b)).
These false matches negatively affect both objectives in our relation prediction model: predicate classification receives incorrect relation labels from non-participating instances, and pair affinity learning (Section~\ref{sec:pals}) suffers from corrupted matched/unmatched partitions where non-interactive pairs are incorrectly treated as positive.

\begin{figure}[t]
    \centering
    \includegraphics[width=1.0\textwidth]{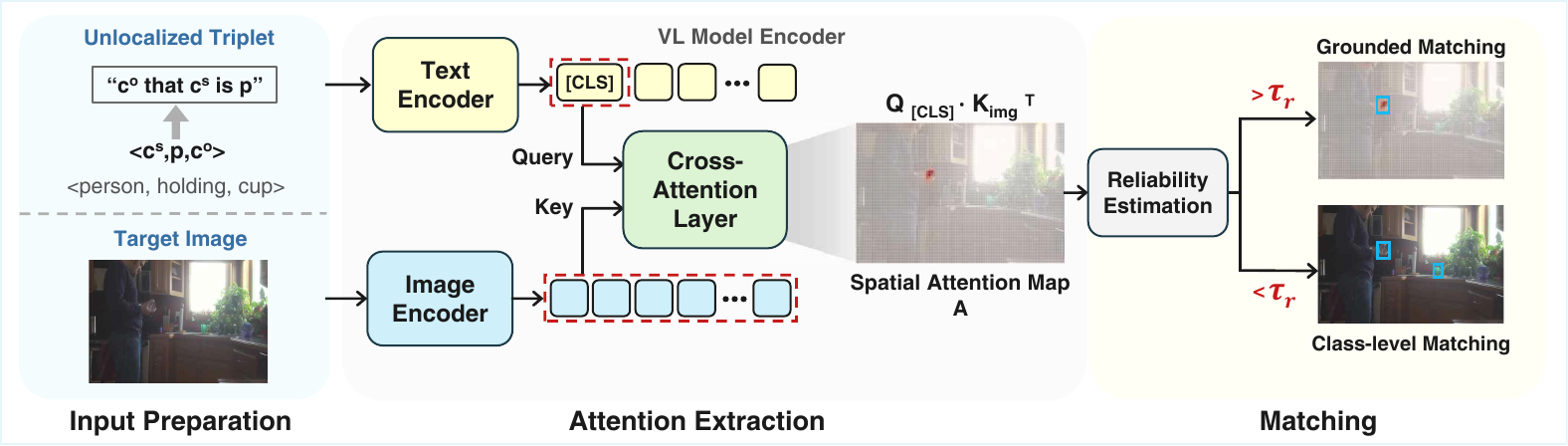}
    \caption{\textbf{Overview of Relation-Aware Matching (RAM).} For each entity in a triplet, a relation-aware query is constructed and fed into a VL grounding model. The \texttt{[CLS]} cross-attention map localizes the described relation, and reliability estimation determines whether to perform grounded matching via Grounding Score (GS) or fall back to class-level matching. The figure illustrates the process for $c^o$; the same procedure is applied independently to $c^s$.}
    \vspace{-0.4cm}
    \label{fig:ram}
\end{figure}

To provide cleaner supervision for both objectives, we refine this partition by leveraging vision-language grounding to identify the most relation-consistent instance among same-class candidates. The key requirement is a mechanism that can associate a relation-aware textual description with a specific object instance, rather than merely identifying objects by category. Vision-language grounding models~\cite{li2022grounded, liu2024grounding} are well suited for this role, as they can localize image regions corresponding to free-form text descriptions. We adopt GroundingDINO~\cite{liu2024grounding} in particular, as its encoder architecture provides readily accessible dense cross-modal attention maps. For each entity in the triplet, we construct a
query that embeds the target class within its relational context (e.g., for $\langle$person, holding, cup$\rangle$, ``cup that the person is holding''). This query is fed into the model along with the input frame, and we extract the \texttt{[CLS]} cross-attention map $\mathbf{A} \in \mathbb{R}^{H_a \times W_a}$ from the final feature enhance layer. Since the query encodes not just the target class but also its relational context, $\mathbf{A}$ naturally concentrates on the specific instance engaged in the described relation.

While attention maps often provide useful spatial signals, they are not always reliable. Since our goal is to identify a single object instance that best matches the described relation, spatially concentrated attention indicates that the model has found a clear visual referent, while dispersed attention suggests either failure to distinguish among candidates or absence of the target object. We leverage this observation by using spatial concentration as a proxy for localization reliability. Given $\mathbf{A}$, we normalize it into a probability distribution $\mathbf{W}$ and define the reliability score $r$ as:
\begin{equation}
\sigma_{\text{spat}} = \sqrt{\sum_{i,j} \mathbf{W}_{ij} \left[ (x_j - \mu_x)^2
+ (y_i - \mu_y)^2 \right]}, \quad
r = \exp\left( -\frac{\sigma_{\text{spat}}}{\sqrt{H_a^2 + W_a^2}} \right),
\label{eq:reliability}
\end{equation}
where $(x_j, y_i)$ are spatial coordinates in the attention map of size $H_a \times W_a$, $(\mu_x, \mu_y)$ is the weighted centroid of $\mathbf{W}$, and $\sigma_{\text{spat}}$ measures spatial dispersion. If $r$ exceeds a threshold $\tau_r$, we perform grounded matching; otherwise, we fall back to class-level matching. For reliable attention maps, each candidate detection $d_k \in \mathcal{D}_{c}$ is scored using a Grounding Score (GS) that combines two complementary measures: concentration $C$, the fraction of total attention mass captured by the detection, and density $\rho$, the attention mass per unit area:
\begin{equation}
C(b_k) = \frac{\sum_{i,j \in b_k} \mathbf{A}_{ij}}{\sum_{i,j} \mathbf{A}_{ij}}, \quad
\rho(b_k) = \frac{\sum_{i,j \in b_k} \mathbf{A}_{ij}}{|b_k|_a},
\end{equation} 
where $|b_k|_a$ denotes the box area in attention-map coordinates. The final grounding score is $\text{GS}(b_k) = C(b_k) \cdot \sigma(\rho(b_k))$, with $\sigma(\cdot)$ denoting the sigmoid function. We select the detection with the highest GS as the match, provided it exceeds a threshold $\tau_{\text{gs}}$. If neither condition is met, the triplet is discarded.

\begin{figure}[t]
    \centering
    \includegraphics[width=0.8\textwidth]{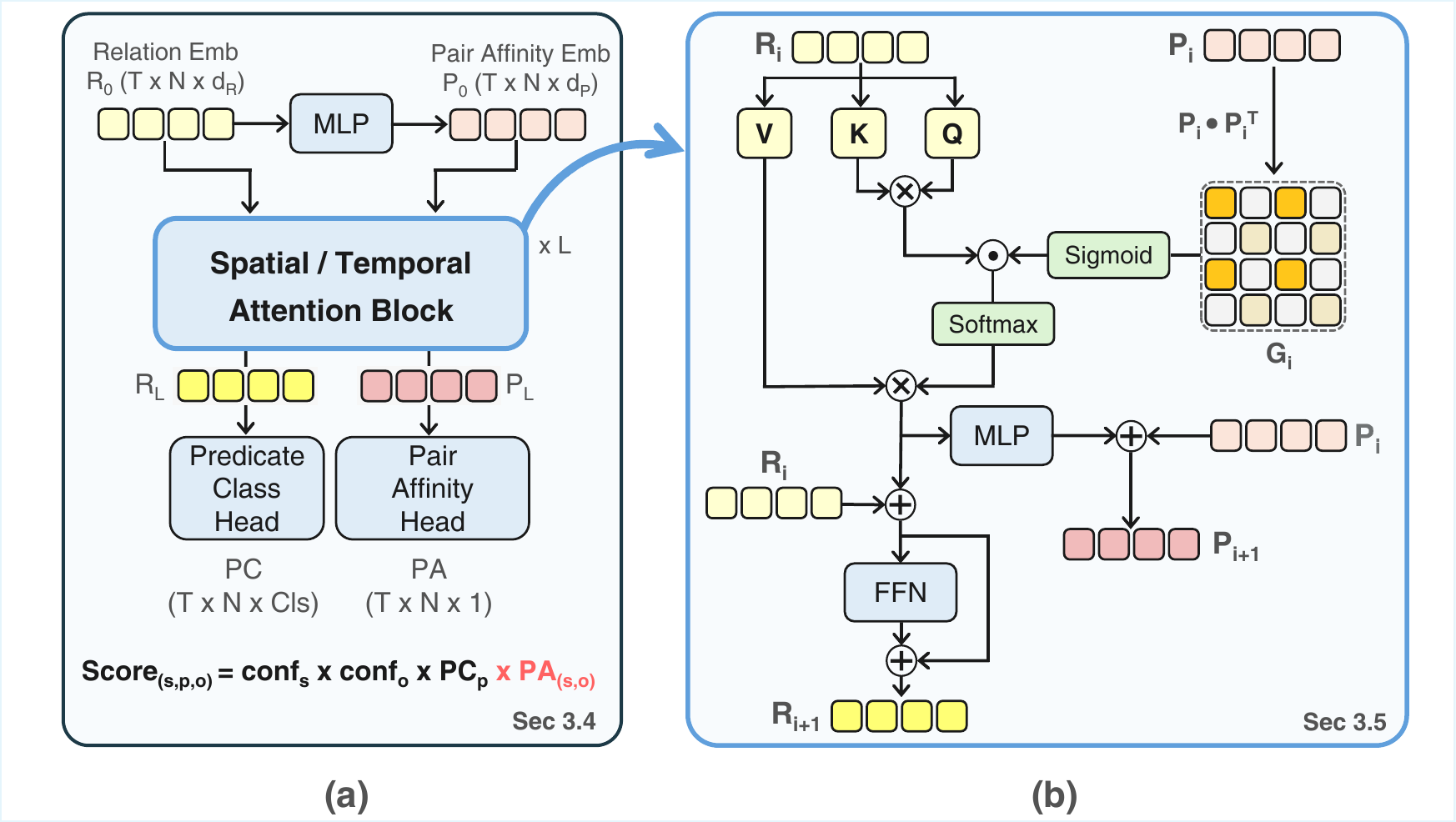} % 확장자 .pdf는 생략 가능
    \caption{Overview of PALS and PAM. \textbf{(a)} The relation embedding $\mathbf{R}_0$ and pair affinity embedding $\mathbf{P}_0$ are jointly updated through $L$ spatial and temporal attention blocks, then decoded into predicate classification scores $\text{PC}$ and pair affinity scores $\text{PA}$ for inference. \textbf{(b)} Inside each attention block, the affinity matrix $\mathbf{G}_i = \mathbf{P}_i \mathbf{P}_i^\top$ gates the attention logits, and the attention output updates both $\mathbf{R}$ and $\mathbf{P}$ via residual connections. The figure illustrates spatial attention for clarity; temporal attention follows the same mechanism with backbone-specific sequence grouping across frames.}
    \label{fig:pals_pam}
    \vspace{-0.4cm}
\end{figure}

\subsection{Pair Affinity Learning and Scoring}
\label{sec:pals}
Given the refined matched/unmatched partition from RAM, the relation prediction model receives all detected pairs. We introduce two parallel embeddings for each pair: a relation embedding $\mathbf{R}$ that encodes which predicate the pair is performing, and a pair affinity embedding $\mathbf{P}$ that encodes whether the pair actively participates in a meaningful interaction. For each subject-object pair, the initial relation embedding $\mathbf{R}_0$ is constructed from the concatenation of subject and object detection features, their spatial union feature, and class embeddings, following the pair representation used in the prior works\cite{cong2021spatial, feng2023exploiting}. The pair affinity embedding $\mathbf{P}_0$ is then derived from $\mathbf{R}_0$ via a projection MLP: $\mathbf{P}_0 = \text{MLP}_{\text{init}}(\mathbf{R}_0)$, as the visual and semantic cues in $\mathbf{R}_0$ already provide a reasonable prior for estimating initial interaction likelihood.

As illustrated in Figure~\ref{fig:pals_pam}(a), both embeddings are jointly refined through $L$ spatial and temporal attention layers, progressively diverging toward their respective objectives. The two embeddings are decoupled since predicate classification and interaction estimation have fundamentally different characteristics: the former is a fine-grained multi-class problem defined over interacting pairs, while the latter is a binary decision that must also handle the far more numerous non-interactive pairs. This distinction is particularly important under weakly-supervised settings, where limited and noisy supervision makes it difficult for a single embedding to implicitly learn both objectives. Explicit separation allows each embedding to be optimized with its own supervision signal and data distribution, reducing the burden on the model under constrained supervision.

After the final layer, the pair affinity score $\text{PA}_{(s,o)}$ is obtained by projecting the pair affinity embedding through an MLP followed by a sigmoid
function:
\begin{equation}
\text{PA}_{(s,o)} = \sigma\!\left(\text{MLP}_{\text{PA}}(\mathbf{P}_L^{(s,o)})\right) \in [0,1],
\end{equation}
\vspace{-0.5cm}
\begin{equation}
\mathcal{L}_{PA} = -\frac{1}{2}\!\left(\frac{1}{|\mathcal{P}^+|}\sum_{(s,o)\in \mathcal{P}^+}\!\log \text{PA}_{(s,o)} + \frac{1}{|\mathcal{P}^-|}\sum_{(s,o)\in \mathcal{P}^-}\!\log(1 - \text{PA}_{(s,o)})\right),
\label{eq:lpa}
\end{equation}
where $\mathcal{P}^+$ and $\mathcal{P}^-$ denote matched and unmatched pairs after RAM, respectively. Since unmatched pairs vastly outnumber matched ones in WS-VSGG, we adopt a class-balanced formulation that averages the loss separately over each set to prevent the majority class from dominating training.

At inference, predicted triplets are ranked by a composite score  to determine the final scene graph. In conventional WS-VSGG pipelines~\cite{chen2022video, xu2025trkt}, this score relies  solely on detection confidence scores $\text{conf}_s$,  $\text{conf}_o$ and predicate classification score $\text{PC}_p$:  $\text{Score}(s, p, o) = \text{conf}_s \cdot \text{conf}_o \cdot  \text{PC}_p$. However, this scoring function can penalize genuinely  interactive pairs. Objects involved in interactions are often partially occluded or motion-blurred, resulting in lower detection confidence that allows clearly visible but non-interactive background objects to outscore them. To explicitly account for interaction likelihood in  the ranking, we incorporate the pair affinity score  $\text{PA}_{(s,o)}$ as an additional multiplicative factor:
\begin{equation}
\text{Score}(s, p, o) = \text{conf}_s \cdot \text{conf}_o \cdot \text{PC}_p \cdot \text{PA}_{(s,o)},
\label{eq:score}
\end{equation}
This ensures that non-interactive pairs are suppressed in the final ranking regardless of their detection confidence or predicate scores, directly addressing the core limitation of existing pipelines.

\subsection{Pair Affinity Modulation}
\label{sec:pam}

While PALS addresses the structural limitation of the training pipeline, the dominance of non-interactive pairs in the detection space also affects the internal reasoning of the relation prediction model. Modern VSGG architectures~\cite{cong2021spatial, feng2023exploiting, nag2023unbiased, khandelwal2024flocode} perform contextual reasoning through attention layers, where each pair attends to other pairs in the sequence to refine its representation. Attending to other interactive pairs provides complementary relational cues beneficial for predicate determination, but the current detection distribution severely limits such interactions in both spatial and temporal attention. In temporal attention, this effect is further compounded as non-interactive pairs from adjacent frames propagate uninformative context across time. To this end, we propose Pair Affinity Modulation (PAM), which modulates attention based on pair affinity embeddings.

Within each attention block at layer $i$, we construct a pair-wise affinity
matrix $\mathbf{G}_i = \mathbf{P}_i \mathbf{P}_i^\top$ over the attention
sequence, where $\mathbf{P}_i$ follows the same sequence composition as
$\mathbf{R}_i$. The composition depends on the attention type: spatial
attention operates over pairs within a single frame, while temporal attention
operates over pairs across multiple frames, with the specific grouping
determined by the backbone
architecture~\cite{cong2021spatial, feng2023exploiting}.
Since $\mathbf{G}_i$ and the attention logits share the same sequence
structure, each element of $\mathbf{G}_i$ directly corresponds to an
attention weight between two pairs, enabling element-wise gating.
We use $\mathbf{G}_i$ to gate the attention logits via element-wise
multiplication with the sigmoid $\sigma(\mathbf{G}_i)$, where
$\mathbf{Q}_i$, $\mathbf{K}_i$, $\mathbf{V}_i$ are the query, key, and
value projections of the relation embedding $\mathbf{R}_i$:
\begin{equation}
\text{Attn}_i =
\text{Softmax}\!\left(
\frac{\mathbf{Q}_i \mathbf{K}_i^\top}{\sqrt{d_k}} \odot \sigma(\mathbf{G}_i)
\right)\mathbf{V}_i,
\label{eq:pam_attn}
\end{equation}
suppressing information flow from low-affinity pairs and redistributing attention mass toward interactive pairs. By filtering noise before it propagates temporally, PAM helps maintain consistent relation predictions across frames. The attention output updates the relation embedding $\mathbf{R}$ via FFN and the pair affinity embedding $\mathbf{P}$ via projection MLP, both with residual connections.
While the gating mechanism modulates attention at every layer, $\mathbf{G}_i$ itself receives no direct supervision. To ensure that the final affinity matrix preserves clear distinction between interactive and non-interactive pairs, we apply triplet ranking loss on $\mathbf{G}_L$:
\begin{equation}
\mathcal{L}_{\text{PAM}} = \sum_{(a,b^+,b^-)} \max\!\left(0,\; \mathbf{G}_L^{(a,b^-)} - \mathbf{G}_L^{(a,b^+)} + m\right),
\label{eq:lpam}
\end{equation}
where $(a, b^+)$ are distinct pairs both matched to annotated relations, $b^-$ is unmatched, and $m$ is a margin. This encourages interactive pairs to attend more strongly to each other than to non-interactive ones, enforcing relative ordering without constraining absolute affinity values.

The overall training objective combines relation classification with pair
affinity losses: $\mathcal{L} = \mathcal{L}_{\text{rel}} + \lambda_{\text{PA}}
\mathcal{L}_{\text{PA}} + \lambda_{\text{PAM}} \mathcal{L}_{\text{PAM}}$,
where $\mathcal{L}_{\text{rel}}$ is the standard relation classification loss applied only to $\mathcal{P}^+$, following the backbone
architecture~\cite{cong2021spatial, feng2023exploiting}. Since PAM operates solely by gating attention logits, it
is applicable as a plug-and-play module to any attention-based relation
prediction architecture, as we verify with both
STTran~\cite{cong2021spatial} and DSG-DETR~\cite{feng2023exploiting} in our
experiments.

\section{Experiments}
\label{sec:exp}

\subsection{Dataset and Implementation Details}
\label{sec:dataset}
We evaluate on Action Genome (AG)~\cite{ji2020action}, which provides frame-level scene graph annotations for 234,253 frames from 9,201 videos, covering 36 object categories and 25 predicate categories. Following prior works~\cite{chen2022video, xu2025trkt, kim2025weakly, peddi2024towards, wang2022dynamic}, we use 7,464 videos for training and 1,737 for testing, with only the unlocalized triplets of the middle frame as supervision. We adopt the Scene Graph Detection (SGDet) protocol, which jointly detects objects and predicts their relations from raw frames without  ground-truth bounding boxes. We report Recall@$K$
($K \in \{10, 20, 50\}$) under two settings: \emph{With Constraint},
where only the top-scoring predicate is retained for each subject-object
pair, and \emph{No Constraint}, where multiple predicates per pair are
permitted.

We follow the two-stage training pipeline of PLA~\cite{chen2022video} and use VinVL~\cite{zhang2021vinvl}, pre-trained on four public object  detection datasets (COCO~\cite{lin2014microsoft}, OpenImages~\cite{kuznetsova2020open}, Objects365~\cite{shao2019objects365}, and Visual Genome~\cite{krishna2017visual}), as the off-the-shelf detector. For the relation prediction backbone, we experiment with both STTran~\cite{cong2021spatial} and DSG-DETR~\cite{feng2023exploiting}. For RAM, we use GroundingDINO~\cite{liu2024grounding} with a Swin-B~\cite{liu2021swin} backbone, with reliability threshold $\tau_r = 0.3$ and grounding score threshold $\tau_{gs} = 0.2$. All experiments are conducted on a single NVIDIA RTX 3090 GPU. Further hyperparameter settings and training
configurations are provided in the supplementary material.

% Requires: \usepackage{multirow}, \usepackage[table]
\begin{table*}[t]
\centering
\caption{Comparison with state-of-the-art methods on the AG dataset (SGDet). Methods are grouped by backbone and supervision level: \emph{Zero-shot} denotes methods without any task-specific training, and \emph{Weak} denotes training with unlocalized scene graph annotations. $^\dagger$ indicates results reproduced with official code under identical settings. Best and second-best results among weakly-supervised methods within each backbone are in \textbf{bold} and \underline{underlined}, respectively.}
\label{tab:sota}
\renewcommand{\arraystretch}{1.15}
\resizebox{\textwidth}{!}{
\begin{tabular}{l|l|l|ccc|ccc}
\toprule
\multirow{2}{*}{Backbone} & \multirow{2}{*}{Supervision} & \multirow{2}{*}{Method} & \multicolumn{3}{c|}{With Constraint} & \multicolumn{3}{c}{No Constraint} \\
 & & & R@10 & R@20 & R@50 & R@10 & R@20 & R@50 \\
\midrule
RLIPv2~\cite{yuan2023rlipv2} & Zero-shot & Vanilla & 5.06 & 8.37 & 10.05 & 5.98 & 14.60 & 21.42 \\
\midrule
\multirow{6}{*}{STTran~\cite{cong2021spatial}}
 & Full & Vanilla                                      & 25.20 & 34.10 & 37.00 & 24.60 & 36.20 & 48.80 \\
\cmidrule(l){2-9}
 & \multirow{5}{*}{Weak}
   & PLA~\cite{chen2022video}                          & 15.39 & 21.44 & 26.24 & 15.83 & 22.83 & 31.74 \\
 & & \cellcolor{gray!15}PLA + \textbf{Ours}           & \cellcolor{gray!15} \textbf{22.24} & \cellcolor{gray!15} \textbf{26.48} & \cellcolor{gray!15} \textbf{28.00} & \cellcolor{gray!15} \textbf{23.20} & \cellcolor{gray!15} \textbf{30.24} & \cellcolor{gray!15} \textbf{37.47} \\
\cmidrule(l){3-9}
 & & TRKT~\cite{xu2025trkt}                            & 17.56 & 22.33 & 27.45 & 18.76 & 24.49 & 33.92 \\
 & & TRKT$^\dagger$                                    & 15.11 & 20.63 & 26.02 & 15.73 & 22.64 & 31.23 \\
 & & \cellcolor{gray!15}TRKT$^\dagger$ + \textbf{Ours} & \cellcolor{gray!15} \underline{19.63} & \cellcolor{gray!15} \underline{24.29} & \cellcolor{gray!15} \underline{27.55} & \cellcolor{gray!15} \underline{21.13} & \cellcolor{gray!15} \underline{27.95} & \cellcolor{gray!15} \underline{35.89} \\
\midrule
\multirow{5}{*}{DSG-DETR~\cite{feng2023exploiting}}
 & Full & Vanilla                                      & 30.30 & 34.80 & 36.10 & 32.10 & 40.90 & 48.30 \\
\cmidrule(l){2-9}
 & \multirow{4}{*}{Weak}
   & PLA                         & 15.47 & 21.33 & 25.86 & 15.66 & 22.71 & 31.90 \\
 & & \cellcolor{gray!15}PLA + \textbf{Ours}           & \cellcolor{gray!15} \textbf{21.88} & \cellcolor{gray!15} \textbf{25.92} & \cellcolor{gray!15} \underline{27.41} & \cellcolor{gray!15} \textbf{23.13} & \cellcolor{gray!15} \textbf{30.34} & \cellcolor{gray!15} \textbf{37.51} \\
\cmidrule(l){3-9}
 & & TRKT$^\dagger$                & 15.23 & 20.10 & 25.92 & 15.60 & 22.09 & 31.18 \\
 & & \cellcolor{gray!15}TRKT$^\dagger$ + \textbf{Ours} & \cellcolor{gray!15} \underline{19.32} & \cellcolor{gray!15} \underline{24.81} & \cellcolor{gray!15} \textbf{27.86} & \cellcolor{gray!15} \underline{21.08} & \cellcolor{gray!15} \underline{28.14} & \cellcolor{gray!15} \underline{35.96} \\
\bottomrule
\end{tabular}
}
\end{table*}

\subsection{Comparison with State-of-the-Art}
\label{sec:sota}

Table~\ref{tab:sota} compares our method with existing approaches on the AG dataset under the SGDet protocol. We apply our full pipeline on top of two WS-VSGG baselines, PLA~\cite{chen2022video} and TRKT~\cite{xu2025trkt}, using two relation prediction backbones, STTran~\cite{cong2021spatial} and DSG-DETR~\cite{feng2023exploiting}. We additionally include RLIPv2~\cite{yuan2023rlipv2}, an image-level relation detection model applied in a zero-shot manner, and the fully-supervised vanilla models as reference points, where the latter serves as an upper bound for each backbone.

Our method yields consistent improvements across all four baseline--backbone combinations, with average gains of +5.47 in R@10 under With Constraint and +6.43 under No Constraint. Improvements are consistent across all K values, as shown in Table~\ref{tab:sota}.
Notably, our best configuration (PLA + Ours on STTran) achieves 88.3\% and 94.3\% of the fully-supervised upper bound at R@10 under the With-Constraint and No-Constraint protocols, respectively, while relying only on single-frame unlocalized annotations without bounding box supervision. These results substantially narrow the gap between weak and full supervision, highlighting the importance of explicitly modeling pairwise interactivity in the weakly-supervised regime.

\subsection{Ablation Study}
\label{sec:ablation}
To analyze the contribution of each component, we conduct an ablation study by incrementally adding RAM, PALS, and PAM to the baseline. We adopt PLA~\cite{chen2022video} as our baseline (row (a)), which uses VinVL~\cite{zhang2021vinvl} as the off-the-shelf detector and STTran~\cite{cong2021spatial} as the relation prediction model. Since PAM modulates attention using the pair affinity embedding that PALS introduces, PAM requires PALS as a prerequisite. When PALS is applied without PAM, the pair affinity score is learned and used only for inference-time ranking, while the attention modulation and the $\mathcal{L}_{\text{PAM}}$ loss are disabled. Table~\ref{tab:ablation} summarizes the results.
\begin{table}[t]
\centering
\caption{Ablation study on the AG dataset (SGDet). Best results are in \textbf{bold} and second best are \underline{underlined}.}
\label{tab:ablation}
\setlength{\tabcolsep}{5.5pt}
\renewcommand{\arraystretch}{0.95}
\footnotesize
\begin{tabular}{c|ccc|ccc|ccc}
\toprule
\multirow{2}{*}{} & \multirow{2}{*}{RAM} & \multirow{2}{*}{PALS} & \multirow{2}{*}{PAM} & \multicolumn{3}{c|}{With Constraint} & \multicolumn{3}{c}{No Constraint} \\
 & & & & R@10 & R@20 & R@50 & R@10 & R@20 & R@50 \\
\midrule
(a) & & & & 15.39 & 21.44 & 26.24 & 15.83 & 22.83 & 31.74 \\
(b) & & \checkmark & & 19.79 & 24.00 & 26.34 & 21.03 & 26.94 & 33.63 \\
(c) & & \checkmark & \checkmark & 20.22 & 24.48 & 26.83 & 21.63 & 27.84 & 34.73 \\
(d) & \checkmark & & & 15.90 & 21.87 & 26.63 & 16.46 & 23.33 & 32.36 \\
(e) & \checkmark & \checkmark & & \underline{22.14} & \underline{25.74} & \underline{27.07} & \underline{23.12} & \underline{29.94} & \underline{36.35} \\
(f) & \checkmark & \checkmark & \checkmark & \textbf{22.24} & \textbf{26.48} & \textbf{28.00} & \textbf{23.20} & \textbf{30.24} & \textbf{37.47} \\
\bottomrule
\end{tabular}
\end{table}

\noindent\textbf{Effect of RAM.}
 RAM alone yields only a modest improvement (row~(a)$\to$(d)); although it reduces label noise, it simultaneously reduces the number of matched pairs while the model still trains exclusively on them. However, RAM substantially amplifies PALS (row~(b)$\to$(e)), demonstrating that its primary value lies in providing cleaner supervision for pair affinity learning.

\noindent\textbf{Effect of PALS.}
Adding PALS yields the largest individual gain (row~(a)$\to$(b)), confirming that the inability to filter non-interactive pairs is the primary bottleneck in WS-VSGG. When combined with RAM (row~(d)$\to$(e)), PALS achieves an even larger gain, as the cleaner matched/unmatched partition directly improves the supervision quality for pair affinity learning.

\noindent\textbf{Effect of PAM.}
PAM provides consistent improvements over PALS alone (row~(b)$\to$(c) and row~(e)$\to$(f)), with gains more pronounced at higher $K$ values, indicating that attention modulation improves predicate classification itself by suppressing noisy context from non-interactive pairs. We further verify this through a controlled experiment in the supplementary material.

\noindent\textbf{Overall.}
The full model (row~(f)) achieves the best performance across all metrics. All three components are synergistic, each yielding progressively larger improvements when the others are present. This interdependence validates our pipeline design, where each stage addresses a distinct weakness of WS-VSGG while reinforcing the others.

\begin{table}[t]
\centering
\caption{Effect of RAM on pseudo-label quality. ``--'' denotes the original matching without RAM. Relative changes are shown in parentheses.}
\label{tab:ram}
\renewcommand{\arraystretch}{1.0}
\footnotesize
\begin{tabular}{l|l|c@{\hspace{8pt}}c|ccc}
\toprule
Detection~ & ~VL Model~ & ~Match~ & ~TP~ & Precision & Recall & F1 \\
\midrule
\multirow{3}{*}{PLA}
 & --    & 16,137 & 6,480 & 0.4016 & \textbf{0.4423} & 0.4209 \\
 & GLIP  & 9,109  & 5,363 & 0.5888 {\scriptsize\textcolor{blue}{(+46.6\%)}} & 0.3660 {\scriptsize\textcolor{red}{($-$17.2\%)}} & 0.4514 {\scriptsize\textcolor{blue}{(+7.2\%)}} \\
 & GDINO & 8,067  & 5,853 & \textbf{0.7255} {\scriptsize\textcolor{blue}{(+80.6\%)}} & 0.3995 {\scriptsize\textcolor{red}{($-$9.7\%)}} & \textbf{0.5153} {\scriptsize\textcolor{blue}{(+22.4\%)}} \\
\midrule
\multirow{3}{*}{TRKT}
 & --    & 21,976 & 7,012 & 0.3191 & \textbf{0.4786} & 0.3829 \\
 & GLIP  & 11,945 & 5,472 & 0.4581 {\scriptsize\textcolor{blue}{(+43.6\%)}} & 0.3735 {\scriptsize\textcolor{red}{($-$22\%)}} & 0.4115 {\scriptsize\textcolor{blue}{(+7.5\%)}} \\
 & GDINO & 10,444 & 5,846 & \textbf{0.5597} {\scriptsize\textcolor{blue}{(+75\%)}} & 0.3990 {\scriptsize\textcolor{red}{($-$17\%)}} & \textbf{0.4659} {\scriptsize\textcolor{blue}{(+22\%)}} \\
\bottomrule
\end{tabular}
\end{table}

\begin{figure}[t]
\centering
\begin{minipage}[b]{0.55\linewidth}
  \centering
  \includegraphics[width=\linewidth]{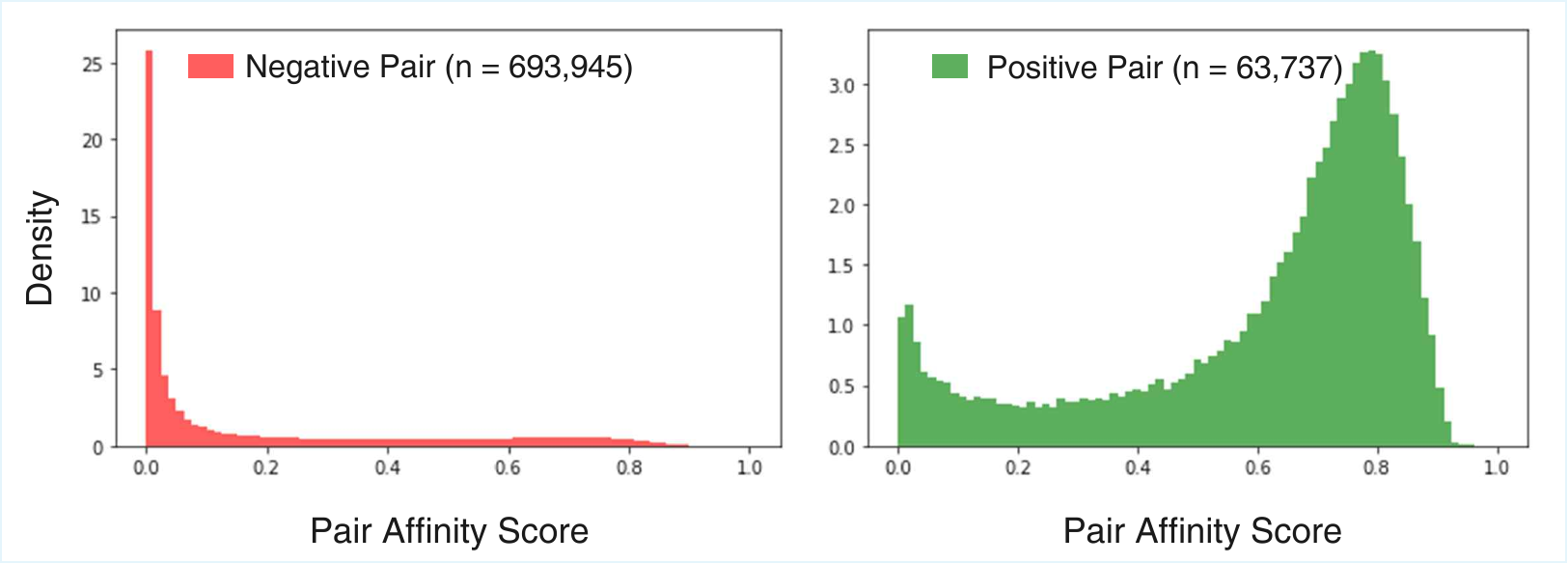}
  \captionof{figure}{Pair affinity score distributions on the AG test set. Negative pairs cluster near zero (left); positive pairs peak around 0.7--0.9 (right).}
  \label{fig:pals_dist}
\end{minipage}%
\hfill
\begin{minipage}[b]{0.43\linewidth}
  \centering
  \renewcommand{\arraystretch}{1.15}
  \scriptsize
  \begin{tabular}{l|ccc|ccc}
  \toprule
   & \multicolumn{3}{c|}{With Constraint} & \multicolumn{3}{c}{No Constraint} \\
  PA & R@10 & R@20 & R@50 & R@10 & R@20 & R@50 \\
  \midrule
  w/o & 17.08 & 23.04 & 27.57 & 16.26 & 23.10 & 32.41 \\
  w/ & \textbf{22.24} & \textbf{26.48} & \textbf{28.00} & \textbf{23.21} & \textbf{30.24} & \textbf{37.47} \\
  \bottomrule
  \end{tabular}
  \vspace{12pt}
  \captionof{table}{Effect of pair affinity scoring at inference. Same weights; only scoring is toggled.}
  \label{tab:pals}
\end{minipage}
\end{figure}

\subsection{Component Analysis}
\label{sec:component}

\noindent\textbf{Pseudo-label refinement via RAM.}
Table~\ref{tab:ram} reports the number of generated pseudo-labels (\textit{Match}), ground-truth true positives (\textit{TP}), and the resulting \textit{Precision}, \textit{Recall}, and \textit{F1} across different detection--VL model combinations. While our main pipeline uses GDINO~\cite{liu2024grounding}, we additionally evaluate with GLIP~\cite{li2022grounded} to verify that RAM is not tied to a specific VL model. RAM consistently improves precision by a large margin regardless of the VL model: GDINO yields +80.6\% for PLA and +75\% for TRKT, while GLIP yields +46.6\% and +43.6\%, respectively, with only modest recall trade-offs. Notably, even after refinement, TRKT's best precision (0.56) remains well below PLA's (0.73), which explains the performance gap between TRKT + Ours and PLA + Ours in Table~\ref{tab:sota}. Further analysis on RAM is provided in the supplementary material.

\noindent\textbf{Effect of pair affinity scoring.}
Fig.~\ref{fig:pals_dist} visualizes the learned pair affinity score distributions on the test set, separated by whether each object pair has a ground-truth relation (Positive, IoU$>$0.5) or not (Negative). The two distributions are clearly separated, confirming that the model effectively learns to distinguish interactive from non-interactive pairs. Table~\ref{tab:pals} further quantifies this effect on end-task performance. Using the trained weights of PLA + Ours with the STTran backbone, toggling pair affinity scoring at inference yields consistent gains across all metrics, with the largest improvements at strict budgets: +5.16 in (W)R@10 and +6.95 in (N)R@10.

\begin{figure*}[t]
\centering
\includegraphics[width=\textwidth]{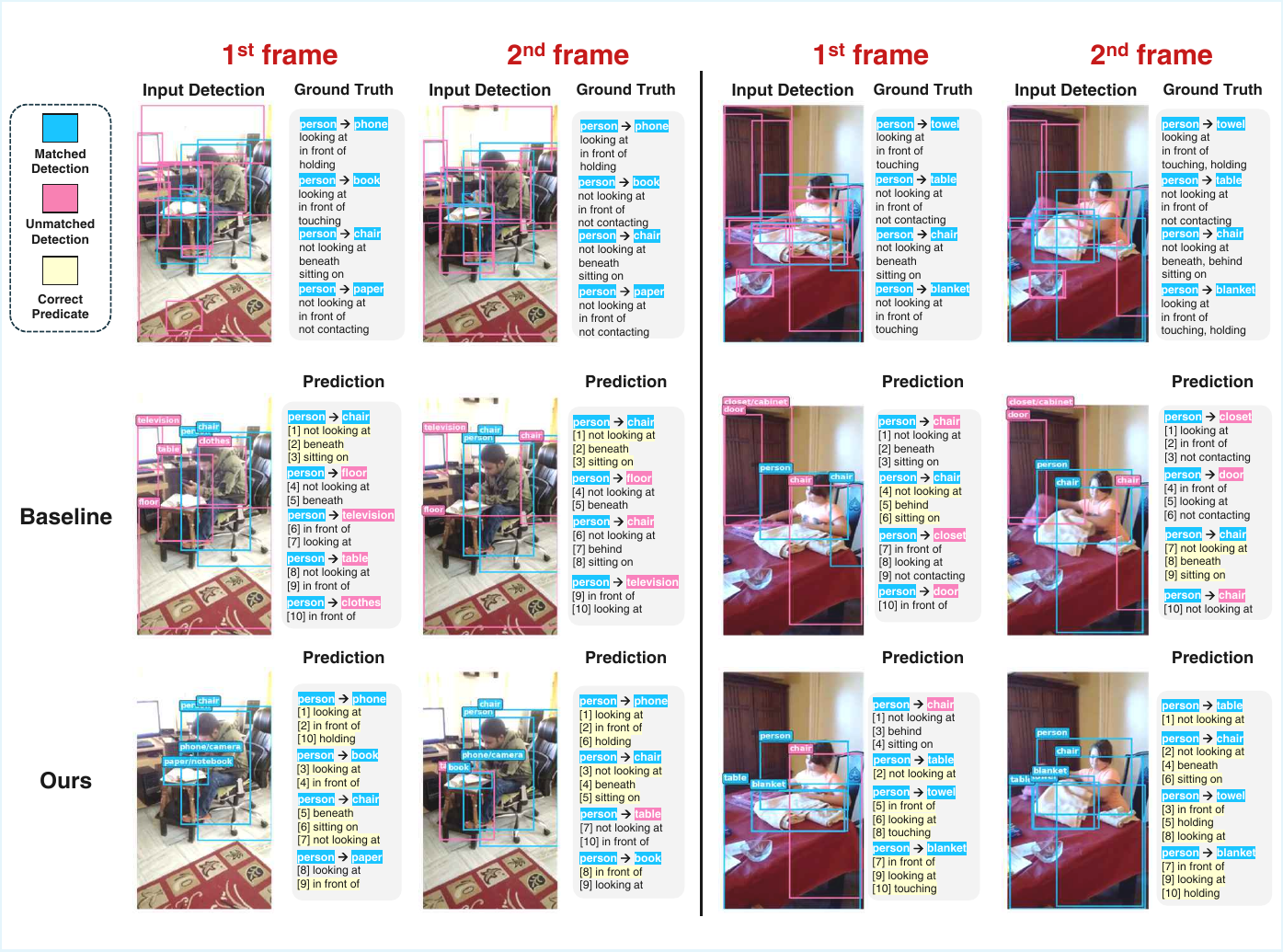}
\caption{Qualitative comparison of top-10 predictions on the AG test set. Two video clips are shown, each with two consecutive frames. For each frame, input detections are overlaid, where ground-truth matched detections are shown in blue and unmatched detections in pink. In the prediction lists, predicates are ranked from [1] (highest score) to [10], and correctly predicted predicates are highlighted in yellow. Compared to the baseline (PLA), our method ranks more interaction-relevant object pairs within the top-10, while the baseline frequently assigns high ranks to non-interactive objects (e.g., floor, clothes, closet, door).}
\vspace{-0.5cm}
\label{fig:qualitative}
\end{figure*}

\subsection{Qualitative Result}
\label{sec:qualitative}

Fig.~\ref{fig:qualitative} presents qualitative comparisons between PLA (baseline) and PLA + Ours on two consecutive frames from the AG test set. For each frame, we overlay the input detections and visualize the top-10 predicted triplets.

Compared to the baseline, our method consistently ranks triplets associated with matched detections within the top-10 predictions. Even in the presence of numerous noisy detections, our model prioritizes objects that participate in annotated interactions. In contrast, the baseline frequently assigns high ranks to unmatched detections, such as \textit{television} and \textit{clothes} in the left example, or \textit{door} and \textit{closet} in the right example. Moreover, our predictions contain more correctly predicted predicates, indicating improved relational accuracy alongside better object prioritization.
These qualitative results demonstrate that our framework effectively suppresses spurious detections and promotes interaction-relevant object pairs, leading to more meaningful scene graph predictions.

\section{Conclusion}
\label{sec:conclusion}
We presented a model-agnostic framework for weakly-supervised video scene graph generation that addresses the core challenge of distinguishing interactive from non-interactive object pairs. RAM refines pseudo-labels via vision-language grounding, PALS learns a pair affinity score to re-rank predictions at inference, and PAM modulates backbone attention to suppress noisy context from non-interactive pairs. Experiments on Action Genome demonstrate consistent improvements across multiple baselines and backbones, narrowing the gap to full supervision to within 88\% and 94\% of the upper bound in (W)R@10 and (N)R@10, respectively. We believe the principle of explicitly modeling interactivity between detected objects can benefit broader weakly-supervised settings beyond scene graph generation.

\vspace{0.7em}
\noindent\textbf{Limitations and future work.}
The quality of pair affinity supervision is inherently bounded by the
matched/unmatched partition. RAM relies on a pretrained VL model, and while
the reliability estimation mitigates uncertain localizations, it remains a
heuristic proxy. Incorrect grounding can still introduce false positives, while
annotation incompleteness may leave false negatives. Developing more principled
approaches to grounding reliability remains an important direction for future
work.

\clearpage  % TODO FINAL: This \clearpage needs to be removed from both review and camera-ready versions.

% \section*{Acknowledgements}

\bibliographystyle{splncs04}
\bibliography{main}

\clearpage
\appendix

\section*{Supplementary Material}
\vspace{0.5cm}
\addcontentsline{toc}{section}{Supplementary Material}

\input{supplementary}

\end{document}

%% file: supplementary.tex
% ============================================================
% ECCV 2026 Supplementary Material Template
% Paper ID: 9410
% ============================================================
%\documentclass[runningheads]{llncs}

% ---- Core Packages ----
%\usepackage{graphicx}
%\usepackage{amsmath,amssymb}
%\usepackage{booktabs}          % 깔끔한 표 (toprule, midrule, bottomrule)
%\usepackage{multirow}          % 표에서 행 병합
%\usepackage{makecell}          % 표 셀 내 줄바꿈
%\usepackage{subcaption}        % subfigure 환경
%\usepackage{xcolor}            % 색상 (highlight 등)
%\definecolor{eccvblue}{rgb}{0.0, 0.3, 0.7}
%\usepackage[pagebackref,breaklinks,colorlinks,citecolor=eccvblue]{hyperref}         % 하이퍼링크
%\usepackage{cleveref}          % \cref 자동 참조 (Fig., Table, Sec. 등)
%\usepackage{algorithm}         % 알고리즘 환경
%\usepackage{algorithmic}       % 알고리즘 pseudo-code
%\usepackage{enumitem}          % 리스트 커스터마이징
%\usepackage{pifont}            % \ding (체크마크 등)
%\usepackage{bm}                % bold math symbols
%\usepackage{colortbl}
% ---- Figure/Table 번호를 본문과 구분 ----
% 본문 Figure 1-6, Table 1-4 이후를 이어서 매기고 싶으면 아래 setcounter 사용
% 별도 번호(S1, S2...)를 쓰고 싶으면 renewcommand 사용
% 방법 1: 본문 이어서 (Figure 7~, Table 5~)
% \setcounter{figure}{6}
% \setcounter{table}{4}
% 방법 2: S 접두사 (Figure S1~, Table S1~) — 아래 주석 해제
% \renewcommand{\thefigure}{S\arabic{figure}}
% \renewcommand{\thetable}{S\arabic{table}}

% ---- Section 번호를 A, B, C로 ----
\renewcommand{\thesection}{\Alph{section}}
\renewcommand{\thesubsection}{\Alph{section}.\arabic{subsection}}

% ---- 유용한 커스텀 명령어 ----
%\newcommand{\ie}{\textit{i.e.}}
%\newcommand{\eg}{\textit{e.g.}}
%\newcommand{\etal}{\textit{et al.}}
%\newcommand{\wrt}{\textit{w.r.t.}}
%\newcommand{\cmark}{\ding{51}}   % ✓
%\newcommand{\xmark}{\ding{55}}   % ✗
%\newcommand{\PA}{\text{PA}}
%\newcommand{\PC}{\text{PC}}

%\usepackage{colortbl}
%\newcommand{\gain}[2]{\shortstack{#1\\\textcolor{red}{\scriptsize(#2)}}}

% ---- 수식 관련 ----
%\DeclareMathOperator*{\argmax}{arg\,max}
%\DeclareMathOperator*{\argmin}{arg\,min}
%\DeclareMathOperator{\sigmoid}{sigmoid}

% ============================================================
%\begin{document}

%\title{Supplementary Material}

%\maketitle
\setcounter{figure}{6}
\setcounter{table}{4} 
\setcounter{equation}{7}
% ============================================================
\section{Framework Details}
\label{sec:implementation}
\vspace{0.2cm}
% ============================================================

\subsection{Training Pipeline}
\label{subsec:training_pipeline}

For completeness, we briefly summarize the two-step training pipeline of PLA~\cite{chen2022video}, which serves as the foundation of our framework.
In the weakly-supervised setting, only the middle frame of each video clip is annotated with unlocalized triplets, providing no bounding box supervision.

\noindent\textbf{Step~1: Pseudo-label generation.}
For the annotated middle frame, detected object proposals are matched to unlocalized triplets based on class identity, producing pseudo-localized scene graph labels.
A relation prediction model (\ie, STTran~\cite{cong2021spatial} or DSG-DETR~\cite{feng2023exploiting}) is then trained on these pseudo labels, referred to as the \emph{model-based teacher}.
To extend supervision beyond the single annotated frame, pseudo labels are propagated to neighboring frames: for each adjacent frame, a detection is adopted as a pseudo label if its bounding box overlaps with the corresponding pseudo ground-truth box from the annotated frame by at least 0.5 IoU.
The relation labels are copied from the annotated frame, as the actual predicates in neighboring frames are unobservable under the weak supervision setting.
These propagated labels constitute the \emph{model-free teacher}.

\noindent\textbf{Step~2: Knowledge distillation.}
In the second step, the model is retrained on all frames using a combination of the propagated pseudo labels (model-free teacher) and the predictions of the model-based teacher from Step~1, combined through an adaptive weighting scheme with a KL divergence loss.
RAM refines the class-level matching in Step~1, while PALS and PAM are incorporated into the relation prediction model in both steps.

\subsection{Adaptations for Knowledge Distillation Step}
\label{subsec:distillation_adaptations}

In Step~1, pair affinity and PAM are supervised using the matched/unmatched partition from the annotated middle frame.
In Step~2, this partition is unavailable for most frames, and the propagated pseudo labels become less reliable with increasing temporal distance.
Label propagation inherently accumulates errors with temporal 
distance, as the IoU-based propagation criterion becomes 
increasingly unreliable for frames far from the annotated 
middle frame.

Inspired by the model-free/model-based blending principle used for relation labels in PLA, we adapt the PA and PAM supervision in a distance-aware manner, where the degree of blending is modulated by the temporal distance from the annotated middle frame.
\clearpage
\noindent\textbf{Distance-aware PA supervision.}
We soften the hard binary PA target by blending it with the model-based teacher's prediction:
\begin{equation}
    y_{(s,o)} = w \cdot y_{(s,o)}^{\text{prop}} + (1 - w) \cdot \text{PA}_{(s,o)}^{\text{teacher}},
\end{equation}
where $y_{(s,o)}^{\text{prop}} \in \{0, 1\}$ is the propagated label, $\text{PA}_{(s,o)}^{\text{teacher}} \in [0,1]$ is the teacher's predicted pair affinity, and $w = 1/(1 + \Delta t)^{\alpha}$ with $\alpha = 3$ decays with temporal distance $\Delta t$.

\noindent\textbf{Adaptive margin for PAM loss.}
We replace the fixed margin $m$ in $\mathcal{L}_{\text{PAM}}$ (Eq.~7) 
with an adaptive margin based on the blended PA target $y_{(s,o)}$ 
defined above. Pairs with $y_{(s,o)} > 0.5$ are treated as positive 
($b^+$) and the rest as negative ($b^-$), with the margin:
\begin{equation}
    m_{(a, b^+, b^-)} = m_{\text{base}} \cdot (y_{b^+} - 0.5) \times 2 
    \cdot (0.5 - y_{b^-}) \times 2.
\end{equation}
This shrinks the margin for uncertain pairs (targets near 0.5) and 
enforces stronger separation for confident ones.

\subsection{Hyperparameters}
\label{subsec:hyperparams}

We use AdamW optimizer with CosineAnnealingLR scheduler across both steps.
Training is performed at the video level, where each video clip constitutes a single batch.
The relation embedding $\mathbf{R}_0 \in \mathbb{R}^{d_R}$ is constructed by concatenating the subject and object visual features (each 512-d), the spatial union feature (512-d), and the subject and object GloVe~\cite{pennington2014glove} word embeddings (each 200-d), yielding $d_R = 1936$.
The pair affinity embedding $\mathbf{P}_0 \in \mathbb{R}^{d_P}$ is derived from $\mathbf{R}_0$ via a 2-layer MLP with LayerNorm and ReLU, projecting to $d_P = 128$.
In the knowledge distillation step, the PA loss uses a soft class-balanced BCE with distance decay $\alpha = 3.0$, and the PAM margin is adaptively scaled based on the teacher's confidence, as described in Section~\ref{subsec:distillation_adaptations}.
Table~\ref{tab:hyperparams} summarizes the hyperparameter settings.
\vspace{-0.5cm}
\begin{table}[h!]
\centering
\caption{Hyperparameter settings.}
\label{tab:hyperparams}
\begin{tabular}{lc}
\toprule
Hyperparameter & Value \\
\midrule
Learning rate & 1e-5 \\
Epochs & 5 \\
Batch size & 1 (video-level) \\
$d_R$ & 1936 \\
$d_P$ & 128 \\
$\lambda_{\text{PA}}$ & 1.0 \\
$\lambda_{\text{PAM}}$ & 0.1 \\
PAM margin $m$ & 1.0 \\
\midrule
Reliability threshold $\tau_r$ & 0.3 \\
Grounding score threshold $\tau_{gs}$ & 0.2 \\
\bottomrule
\end{tabular}
\end{table}
\vspace{-0.7cm}

\subsection{Computation Analysis}
\label{subsec:computation}

We measure the computational overhead introduced by PALS and PAM on top of both backbone architectures.
For FLOPs, we use the average input size of the Action Genome test set ($T\!=\!30$ frames, $N\!=\!18$ pairs per frame based on VinVL~\cite{zhang2021vinvl} detections).
As shown in Table~\ref{tab:computation}, PALS and PAM add only 1.32M parameters (+1.27\%) with negligible FLOPs and latency overhead on both backbones.
RAM is a one-time preprocessing step applied only during training and incurs no overhead at inference.

\begin{table}[h!]
\centering
\caption{Computational overhead of PALS and PAM. FLOPs are 
measured with $T\!=\!30$, $N\!=\!18$. Latency is measured 
under the same setting and averaged over 100 runs on a 
single NVIDIA RTX 3090 GPU.}
\label{tab:computation}
\setlength{\tabcolsep}{7.5pt}
\begin{tabular}{lccc}
\toprule
Model & Params (M) & FLOPs (G) & Latency (ms/video) \\
\midrule
STTran & 103.67 & 138.91 & 20.95 \\
STTran + Ours & 104.99 {\color{red}\scriptsize(+1.27\%)} & 138.92 {\color{red}\scriptsize(+0.01\%)} & 20.95 {\color{red}\scriptsize(+0\%)} \\
\midrule
DSG-DETR & 103.67 & 140.05 & 18.09 \\
DSG-DETR + Ours & 104.99 {\color{red}\scriptsize(+1.27\%)} & 140.10 {\color{red}\scriptsize(+0.04\%)} & 18.10 {\color{red}\scriptsize(+0.05\%)} \\
\bottomrule
\end{tabular}
\end{table}
\vspace{-0.7cm}
% ============================================================
\section{Experiments on NL-VSGG Setting}
\label{sec:nlvsgg}
% ============================================================
% Caption-based weak supervision에서의 실험 결과

\subsubsection{Generalization to Different Weak Supervision Types}
To verify that our framework generalizes beyond unlocalized triplet 
supervision, we apply our pipeline on top of NL-VSGG~\cite{kim2025weakly}, 
which constructs pseudo-localized training pairs from video captions rather than manually annotated unlocalized triplets, representing a 
strictly weaker form of supervision. As shown in 
Table~\ref{tab:nlvsgg}, our method consistently improves NL-VSGG 
across both STTran and DSG-DETR backbones, demonstrating that our 
framework is agnostic to the source of weak supervision and remains 
effective even when initial pseudo-labels are derived from 
automatically collected captions.

\begin{table}[h]
\centering
\caption{Performance of our method applied on top of NL-VSGG on the 
AG dataset (SGDet).}
\vspace{0.2cm}
\label{tab:nlvsgg}
\setlength{\tabcolsep}{5pt}
\begin{tabular}{l|l|ccc|ccc}
\toprule
\multirow{2}{*}{Backbone} & \multirow{2}{*}{Method} 
& \multicolumn{3}{c|}{With Constraint} 
& \multicolumn{3}{c}{No Constraint} \\
& & R@10 & R@20 & R@50 & R@10 & R@20 & R@50 \\
\midrule
\multirow{2}{*}{STTran}
& NL-VSGG                                        
& 9.48  & 15.61 & 19.60 & 9.48  & 15.92 & 22.56 \\
& \cellcolor{gray!15}NL-VSGG + Ours    
& \cellcolor{gray!15}\textbf{13.42} 
& \cellcolor{gray!15}\textbf{19.77} 
& \cellcolor{gray!15}\textbf{20.51} 
& \cellcolor{gray!15}\textbf{13.44} 
& \cellcolor{gray!15}\textbf{20.60} 
& \cellcolor{gray!15}\textbf{26.31} \\
\midrule
\multirow{2}{*}{DSG-DETR}
& NL-VSGG                                        
& 9.51  & 15.75 & 20.40 & 9.53  & 16.11 & 23.21 \\
& \cellcolor{gray!15}NL-VSGG + Ours    
& \cellcolor{gray!15}\textbf{13.45} 
& \cellcolor{gray!15}\textbf{19.63} 
& \cellcolor{gray!15}\textbf{20.98} 
& \cellcolor{gray!15}\textbf{13.51} 
& \cellcolor{gray!15}\textbf{20.31} 
& \cellcolor{gray!15}\textbf{26.82} \\
\bottomrule
\end{tabular}
\end{table}

% ============================================================
\section{Additional Analysis on RAM}
\label{sec:ram_analysis}
% ============================================================

\subsection{Threshold Sensitivity}
\label{subsec:threshold_sensitivity}

\begin{figure}[h!]
\centering
\begin{subfigure}[t]{\linewidth}
    \centering
    \includegraphics[width=\linewidth]{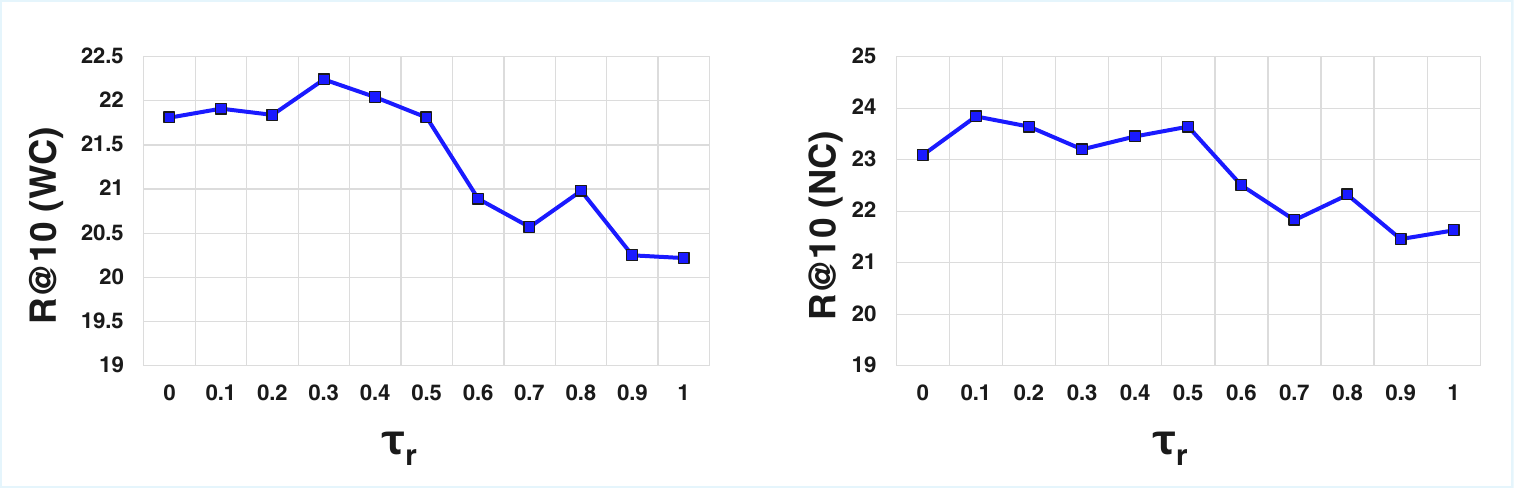}
    \caption{$\tau_r$ sweep ($\tau_{gs}\!=\!0.2$ fixed).}
    \label{fig:sensitivity_taur}
\end{subfigure}
\\[0.5em]
\begin{subfigure}[t]{\linewidth}
    \centering
    \includegraphics[width=\linewidth]{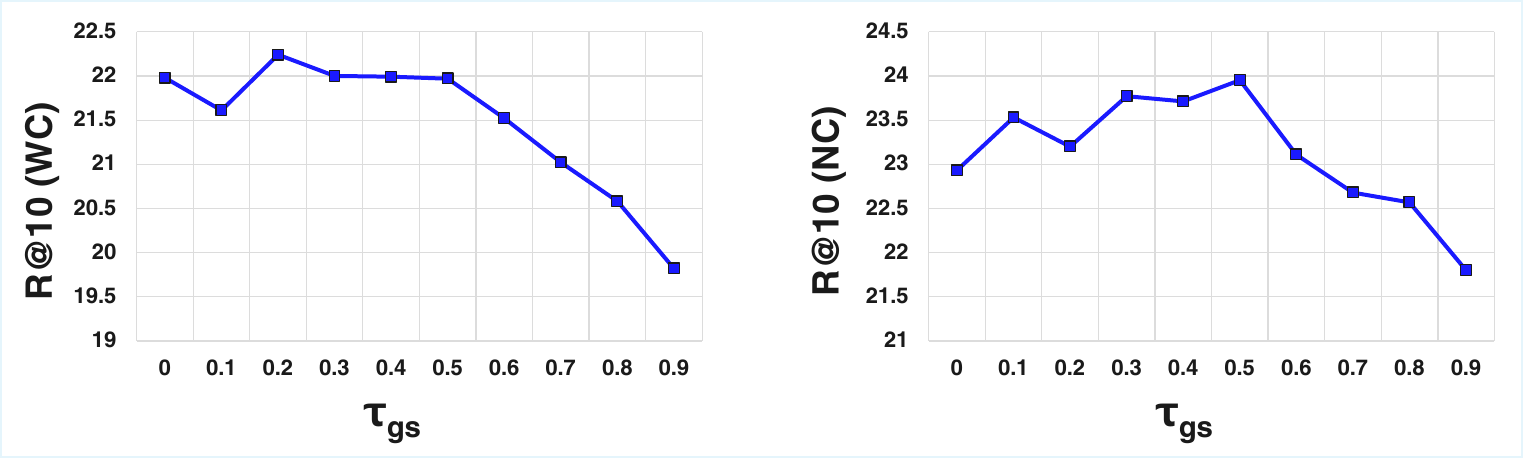}
    \caption{$\tau_{gs}$ sweep ($\tau_r\!=\!0.3$ fixed).}
    \label{fig:sensitivity_taugs}
\end{subfigure}
\caption{Sensitivity of R@10 to RAM thresholds on AG (SGDet). WC and NC denote With Constraint and No Constraint, respectively.}
\label{fig:sensitivity}
\end{figure}

We analyze the sensitivity of RAM to the reliability threshold $\tau_r$ and the grounding score threshold $\tau_{gs}$ by fixing one at its default value and sweeping the other from 0.0 to 1.0.
All experiments use PLA with STTran on the AG dataset.
Figure~\ref{fig:sensitivity} shows R@10 under both protocols, and Tables~\ref{tab:sensitivity_taur} and~\ref{tab:sensitivity_taugs} provide full results including pseudo-label quality.

As shown in Figure~\ref{fig:sensitivity}, performance remains stable across a wide range for both thresholds ($\tau_r \in [0.1, 0.5]$ and $\tau_{gs} \in [0.0, 0.5]$), confirming that the method is not sensitive to the specific choice of these values.
At the extremes, performance degrades as expected: $\tau_r\!=\!0.0$ introduces noise from unreliable attention maps, while high $\tau_{gs}$ discards too many matches, reducing training data.
Notably, even at $\tau_r\!=\!1.0$, where RAM degenerates to class-level matching, the model still substantially outperforms the baseline (W-R@10: 20.25 vs.\ 15.39), confirming the independent contribution of PALS and PAM.

\begin{table}[h!]
\centering
\caption{Sensitivity to $\tau_r$ ($\tau_{gs}\!=\!0.2$ fixed). The shaded row indicates the default setting.}
\label{tab:sensitivity_taur}
\setlength{\tabcolsep}{6pt}
\begin{tabular}{c|ccc|ccc|ccc}
\toprule
\multirow{2}{*}{$\tau_r$} & \multicolumn{3}{c|}{Pseudo-label} & \multicolumn{3}{c|}{With Constraint} & \multicolumn{3}{c}{No Constraint} \\
& Precision & Recall & F1 & R@10 & R@20 & R@50 & R@10 & R@20 & R@50 \\
\midrule
0.0 & 0.694 & 0.372 & 0.485 & 21.81 & 25.55 & 26.74 & 23.09 & 30.00 & 35.92 \\
0.1 & 0.718 & 0.389 & 0.504 & 21.91 & 25.48 & 26.93 & 23.84 & 30.39 & 36.13 \\
0.2 & 0.726 & 0.399 & 0.515 & 21.84 & 25.63 & 27.11 & 23.64 & 30.30 & 37.14 \\
\rowcolor{gray!15} 0.3 & 0.726 & 0.400 & 0.515 & 22.24 & 26.48 & 28.00 & 23.20 & 30.24 & 37.47 \\
0.4 & 0.701 & 0.402 & 0.511 & 22.00 & 25.82 & 27.47 & 23.45 & 29.72 & 36.15 \\
0.5 & 0.641 & 0.408 & 0.498 & 21.81 & 25.55 & 27.01 & 23.64 & 30.21 & 36.93 \\
0.6 & 0.567 & 0.414 & 0.478 & 20.89 & 25.03 & 26.78 & 22.50 & 29.21 & 36.15 \\
0.7 & 0.508 & 0.422 & 0.461 & 20.57 & 24.96 & 27.07 & 21.83 & 28.59 & 35.73 \\
0.8 & 0.437 & 0.435 & 0.436 & 20.98 & 25.09 & 26.90 & 22.32 & 28.79 & 35.75 \\
0.9 & 0.412 & 0.441 & 0.426 & 20.25 & 24.88 & 27.00 & 21.46 & 28.23 & 34.99 \\
1.0 & 0.402 & 0.442 & 0.421 & 20.25 & 24.48 & 26.83 & 21.63 & 27.84 & 34.73 \\
\bottomrule
\end{tabular}
\end{table}

\begin{table}[h!]
\centering
\caption{Sensitivity to $\tau_{gs}$ ($\tau_r\!=\!0.3$ fixed). The shaded row indicates the default setting.}
\label{tab:sensitivity_taugs}
\setlength{\tabcolsep}{6pt}
\begin{tabular}{c|ccc|ccc|ccc}
\toprule
\multirow{2}{*}{$\tau_{gs}$} & \multicolumn{3}{c|}{Pseudo-label} & \multicolumn{3}{c|}{With Constraint} & \multicolumn{3}{c}{No Constraint} \\
& Precision & Recall & F1 & R@10 & R@20 & R@50 & R@10 & R@20 & R@50 \\
\midrule
0.0 & 0.667 & 0.410 & 0.507 & 21.98 & 26.18 & 27.75 & 22.93 & 29.89 & 37.22 \\
0.1 & 0.711 & 0.401 & 0.513 & 21.61 & 25.32 & 26.88 & 23.53 & 30.25 & 37.05 \\
\rowcolor{gray!15} 0.2 & 0.726 & 0.400 & 0.515 & 22.24 & 26.48 & 28.00 & 23.20 & 30.24 & 37.47 \\
0.3 & 0.736 & 0.392 & 0.512 & 22.00 & 25.79 & 27.36 & 23.77 & 30.29 & 36.48 \\
0.4 & 0.747 & 0.383 & 0.506 & 21.99 & 25.61 & 27.12 & 23.71 & 30.10 & 36.51 \\
0.5 & 0.762 & 0.366 & 0.494 & 21.97 & 25.52 & 27.14 & 23.95 & 30.40 & 36.92 \\
0.6 & 0.777 & 0.342 & 0.475 & 21.52 & 25.12 & 26.77 & 23.11 & 29.66 & 36.37 \\
0.7 & 0.790 & 0.301 & 0.436 & 21.02 & 24.66 & 26.46 & 22.68 & 29.01 & 35.71 \\
0.8 & 0.801 & 0.245 & 0.376 & 20.58 & 24.30 & 26.20 & 22.57 & 28.78 & 35.15 \\
0.9 & 0.830 & 0.163 & 0.272 & 19.82 & 23.13 & 25.03 & 21.80 & 27.75 & 34.25 \\
\bottomrule
\end{tabular}
\end{table}

\subsection{Comparison of VL Models}
\label{subsec:vl_models}

We investigate the applicability of our framework across a range of 
vision-language (VL) models to identify the architectural conditions 
necessary for language-guided localization. Specifically, we evaluate 
five models: GroundingDINO~\cite{liu2024grounding}, GLIP~\cite{li2022blip}, 
MDETR~\cite{kamath2021mdetr}, SigLIP2~\cite{tschannen2025siglip}, and BLIP~\cite{li2022blip}.

\subsubsection{Models with explicit region-text alignment.}
\begin{figure}[h]
    \centering
    \includegraphics[width=\linewidth]{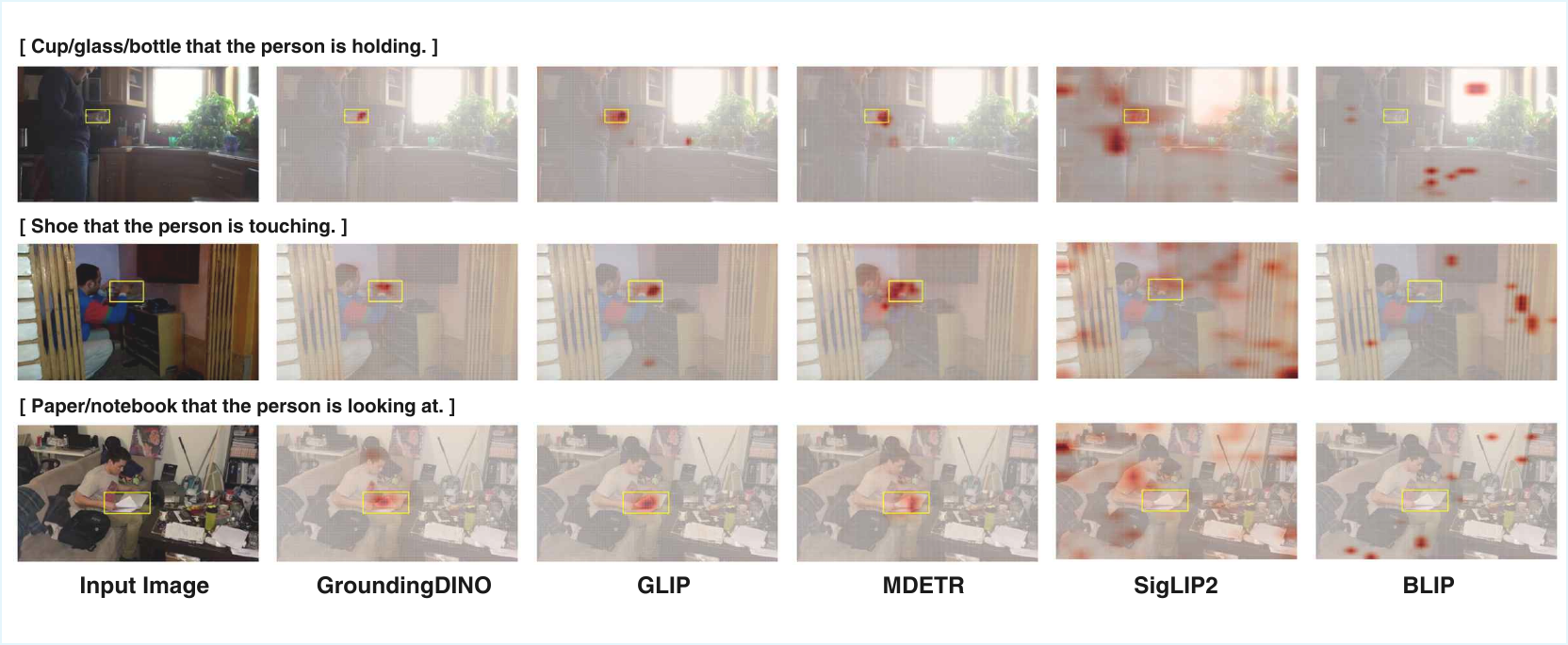}
    \caption{Attention map comparison across five vision-language 
    models on three multi-instance scenarios. Each row shows the same 
    input frame with a language query (shown in brackets) that requires 
    relational understanding to identify the correct target among 
    multiple instances of the same category. Yellow boxes indicate 
    ground-truth regions. GroundingDINO, GLIP, and MDETR produce 
    activations consistently concentrated on the query-relevant region, 
    whereas SigLIP2 and BLIP fail to localize the target, exhibiting 
    diffuse or query-agnostic activation patterns.}
    \vspace{-0.5cm}
    \label{fig:vl_comparison}
\end{figure}

GroundingDINO, GLIP, and MDETR are trained with region-text matching 
objectives that explicitly align language tokens to spatial regions in 
the image. Given a language query, these models produce localized 
predictions grounded in the input text, enabling precise identification 
of query-relevant regions. As shown in Fig.~\ref{fig:vl_comparison}, 
all three models successfully produce attention activations 
concentrated on the target region across three multi-instance 
scenarios, where the correct object must be identified among multiple 
instances of the same category based on relational expressions 
(\emph{e.g.}, ``cup that the person is holding'', ``shoe that the 
person is touching'').
\vspace{-0.5cm}

To further validate compatibility, we apply our full pipeline to all 
three models under the same hyperparameter setting 
($\tau_r\!=\!0.3$, $\tau_{gs}\!=\!0.2$) and report pseudo-label 
quality and retrieval performance in Table~\ref{tab:vl_model_comp}. 
All three models yield comparable retrieval performance, particularly 
under the No Constraint setting (R@50: 37.47, 36.80, 37.20 for 
GroundingDINO, GLIP, and MDETR, respectively), confirming that 
our framework is not specific to a single backbone. 
Among the three, GroundingDINO achieves the highest pseudo-label 
quality (F1: 0.515) and With Constraint retrieval performance 
(R@50: 28.00), and we therefore adopt it as the primary backbone 
in subsequent experiments.

\begin{table}[h!]
\centering
\caption{Performance comparison of compatible VL models 
under the same hyperparameter setting  
($\tau_r\!=\!0.3$, $\tau_{gs}\!=\!0.2$). \textbf{Bold} indicates 
the best result.}
\vspace{0.2cm}
\label{tab:vl_model_comp}
\setlength{\tabcolsep}{4pt}
\begin{tabular}{l|ccc|ccc|ccc}
\toprule
\multirow{2}{*}{Model} & \multicolumn{3}{c|}{Pseudo-label} 
& \multicolumn{3}{c|}{With Constraint} 
& \multicolumn{3}{c}{No Constraint} \\
& Prec. & Rec. & F1 & R@10 & R@20 & R@50 
& R@10 & R@20 & R@50 \\
\midrule
GLIP   & 0.589 & 0.366 & 0.451 & 21.21 & 25.10 & 26.65 
       & 22.86 & 29.66 & 36.80 \\
MDETR  & \textbf{0.730} & 0.369 & 0.490 & 22.11 & 25.66 & 27.10 
       & 22.96 & 30.18 & 37.20 \\
\rowcolor{gray!15}
GroundingDINO & 0.726 & \textbf{0.400} & \textbf{0.515} & \textbf{22.24} & \textbf{26.48} & \textbf{28.00} 
              & \textbf{23.20} & \textbf{30.24} & \textbf{37.47} \\
\bottomrule
\end{tabular}
\end{table}

\subsubsection{Models without explicit region-text alignment.}
We also examine two models that incorporate vision-language interaction 
but lack explicit object-level grounding supervision.

BLIP~\cite{li2022blip} employs cross-attention between visual and language 
modalities, which might suggest local alignment capability. However, 
its training objective focuses on global image-text alignment 
(\emph{e.g.}, image-text contrastive loss, image-text matching), 
rather than aligning individual language tokens to specific image 
regions. As a result, BLIP produces sparse, query-agnostic activations 
that fail to localize the target object 
(Fig.~\ref{fig:vl_comparison}, rightmost column).

SigLIP2~\cite{tschannen2025siglip} explicitly improves 
localization and dense prediction capabilities over the 
original SigLIP through self-supervised objectives, and 
we attempt to exploit this by computing cosine similarity 
between patch-level visual features and text query 
embeddings as a grounding proxy. However, as this 
localization capability is a by-product of self-supervised 
pre-training rather than direct grounding supervision, 
the resulting similarity maps are diffuse and inconsistent, 
failing to reliably identify the query-relevant object 
(Fig.~\ref{fig:vl_comparison}, second column from right).

\subsubsection{Summary.}
These results suggest that the critical requirement for applying our 
framework is not merely the presence of vision-language interaction 
(\emph{e.g.}, cross-attention in BLIP) or improved spatial awareness 
through self-supervised pre-training (\emph{e.g.}, 
SigLIP2), but 
rather explicit region-text alignment learned through grounding 
supervision. Among the evaluated models, GroundingDINO, GLIP, and 
MDETR satisfy this condition and produce consistent results across 
our pipeline, demonstrating the generalizability of our approach 
beyond a single VL backbone.

\begin{figure}[h]
    \centering
    \includegraphics[width=\linewidth]{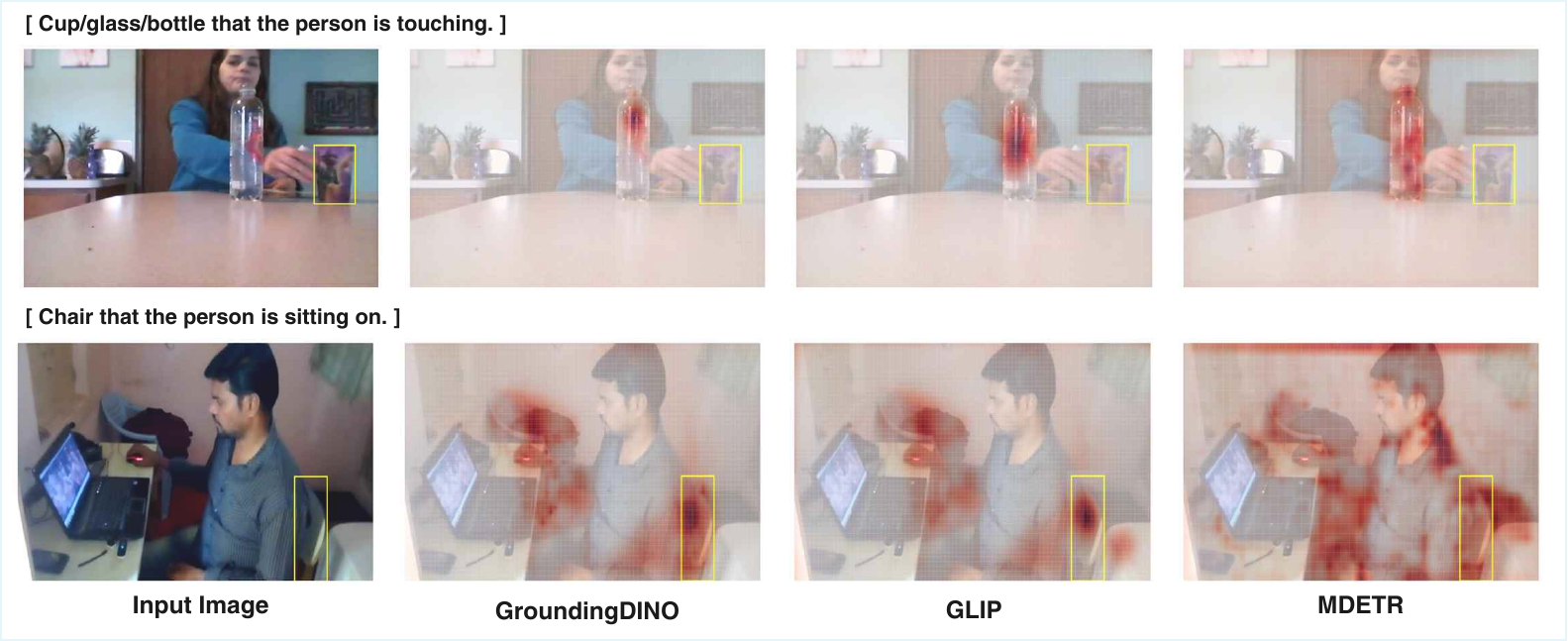}
    \caption{Failure cases of RAM across GroundingDINO, 
    GLIP, and MDETR. Yellow boxes indicate ground-truth 
    regions. Top: categorical ambiguity of the 
    relation-satisfying object. Bottom: heavy occlusion 
    of the relation-satisfying object.}
    \label{fig:failure_cases}
\end{figure}

\subsection{Failure Case Analysis}
\label{subsec:failure_cases}

While RAM with GroundingDINO applied to PLA detection 
significantly improves pseudo-label precision, it 
also introduces a trade-off. Compared to 
class-level matching, RAM improves label accuracy 
by 80\%, but incurs a loss of 9.7\% in true positive 
matches. This corresponds to 627 out of 14,652 
annotations (4.3\%) where a pseudo-label is assigned 
to an incorrect object despite a reliability score 
exceeding $\tau_r$.

To understand the underlying causes, we analyze 
these failure cases and identify a common pattern 
across all three compatible VL models 
(GroundingDINO, GLIP, and MDETR), as illustrated 
in Fig.~\ref{fig:failure_cases}. The failures share 
a common root: when the object satisfying the 
relational condition cannot be confirmed 
categorically with sufficient confidence, the 
grounding model substitutes it with a more 
visually unambiguous candidate of the same 
category, even if that candidate does not satisfy 
the relational condition.

This failure mode arises whenever the grounding
model lacks sufficient visual evidence to
categorically confirm the relation-satisfying
object, leading it to substitute a more visually
unambiguous candidate instead. Such situations
commonly occur when the target object is
visually ambiguous in appearance or heavily
occluded, leaving its categorical identity
unverifiable. As shown in
Fig.~\ref{fig:failure_cases}, both scenarios
result in the model redirecting attention to a
categorically unambiguous but relationally
incorrect instance, and this pattern is
consistently observed across all three VL models,
suggesting that it reflects a general
characteristic of vision-language grounding
models rather than an artifact of a particular
architecture.

% ============================================================
\section{Additional Analysis on PALS}
\label{sec:pals_analysis}
% ============================================================

\subsection{Effect of Loss Formulation for Pair Affinity Learning}
\label{subsec:loss_formulation}

Since unmatched pairs vastly outnumber matched ones in WS-VSGG 
(693,945 vs. 63,737; ratio $\approx$11:1), the choice of loss 
formulation for $\mathcal{L}_{PA}$ has a direct impact on the 
quality of the learned pair affinity scores. We compare standard 
BCE against our class-balanced BCE, which averages the loss 
separately over $\mathcal{P}^+$ and $\mathcal{P}^-$ to prevent 
the majority class from dominating training.

\begin{figure}[h]
    \centering
    \includegraphics[width=\linewidth]{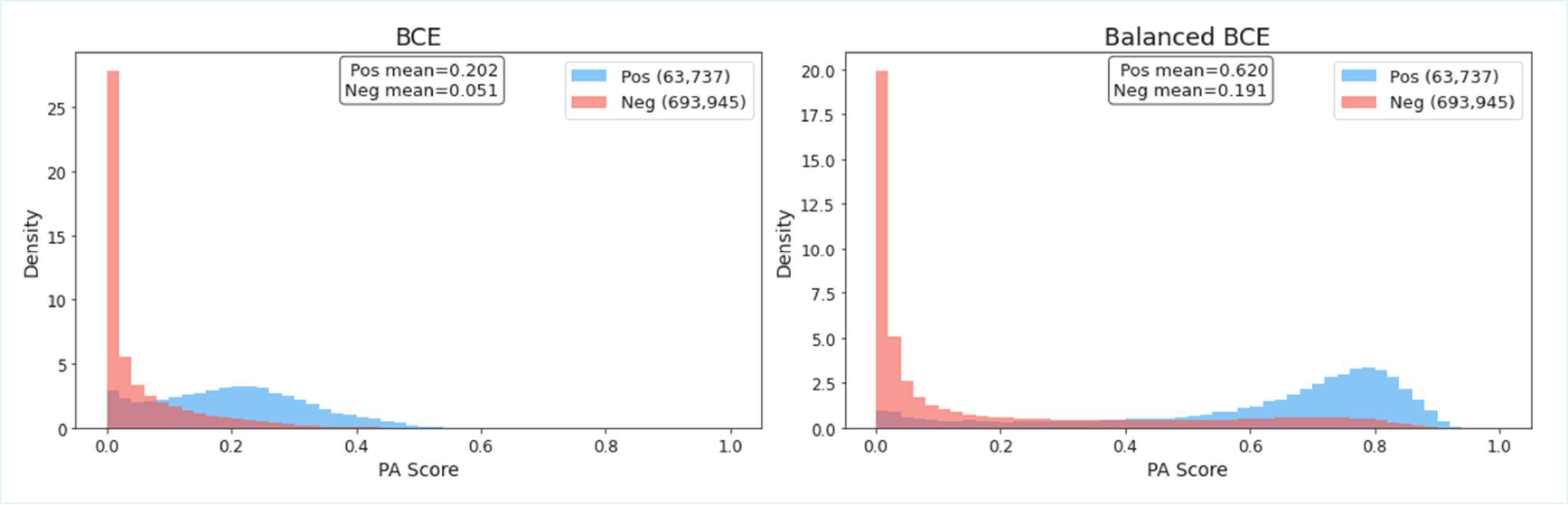}
    \caption{PA score distributions under standard BCE (left) and 
    class-balanced BCE (right). Standard BCE collapses both 
    distributions toward zero due to class imbalance, whereas 
    class-balanced BCE produces well-separated positive and negative 
    distributions.}
    \label{fig:pa_bce}
\end{figure}

As shown in Fig.~\ref{fig:pa_bce}, standard BCE causes the model 
to collapse toward the majority class: both positive and negative 
pairs cluster near zero (Pos mean: 0.202, Neg mean: 0.051), with 
heavily overlapping distributions that provide little discriminative 
signal for ranking. In contrast, class-balanced BCE produces 
well-separated distributions (Pos mean: 0.620, Neg mean: 0.191), 
consistent with the score distributions reported in the main paper 
(Fig. 5). This separation directly translates to improved retrieval 
performance, as shown in Table~\ref{tab:loss_formulation}: 
class-balanced BCE yields consistent gains across all metrics 
(e.g., +0.30 in (W)R@10, +0.88 in (W)R@50, +1.04 in (N)R@10, 
+0.61 in (N)R@50), confirming that proper handling of class 
imbalance is essential for effective pair affinity learning.

\vspace{-1.5cm}
\begin{table}[h]
\centering
\caption{Effect of loss formulation on pair affinity learning.}
\label{tab:loss_formulation}
\setlength{\tabcolsep}{7pt}
\begin{tabular}{l|ccc|ccc}
\toprule
Loss & \multicolumn{3}{c|}{With Constraint} 
     & \multicolumn{3}{c}{No Constraint} \\
     & R@10 & R@20 & R@50 & R@10 & R@20 & R@50 \\
\midrule
BCE              & 21.94 & 25.60 & 27.10 
                 & 22.16 & 30.17 & 36.84 \\
\rowcolor{gray!15}
Balanced BCE (default) & \textbf{22.24} & \textbf{26.48} 
                    & \textbf{28.00} & \textbf{23.20} 
                    & \textbf{30.24} & \textbf{37.47} \\
\bottomrule
\end{tabular}
\end{table}

% ============================================================
\section{Additional Analysis on PAM}
\label{sec:pam_analysis}
% ============================================================
\subsection{Effect by Non-interactive Pair Ratio}
\label{subsubsec:neg_ratio}

To analyze whether PAM's benefit is modulated by the prevalence 
of non-interactive pairs, we partition the AG test set into two 
subsets based on the per-video ratio of non-interactive pairs: 
the top 25\% of videos with the highest ratio (High NI; 434 videos, 
non-interactive ratio 0.928--0.993) and the bottom 25\% with the 
lowest ratio (Low NI; 434 videos, non-interactive ratio 
0.527--0.895), where the ratio is computed by aggregating pair 
counts across all frames in each video.

As shown in Table~\ref{tab:neg_ratio}, two observations emerge. 
First, absolute performance is substantially lower in the High NI 
subset than in the Low NI subset across all configurations, 
confirming that a higher prevalence of non-interactive pairs 
makes relation prediction more challenging. Second, PAM yields a 
markedly larger improvement in the High NI subset than in the 
Low NI subset (e.g., $+4.28$ vs. $+0.38$ in (W)R@10), 
demonstrating that PAM is particularly effective at mitigating 
the adverse effects of non-interactive pairs on contextual 
reasoning. Notably, even in the Low NI subset where non-interactive 
pairs are less prevalent, PAM still yields consistent 
gains across all metrics, suggesting that suppressing 
the attention contribution of non-interactive pairs 
through gating remains effective even when their 
proportion is relatively low.

\begin{table}[h]
\centering
\caption{Performance on video subsets grouped by non-interactive 
pair ratio. High NI and Low NI denote the top and bottom 25\% of 
test videos ranked by per-video non-interactive pair ratio, 
respectively. The w/o PAM model corresponds to row~(e) of 
Table~2 in the main paper (RAM + PALS, without PAM). 
\textcolor{red}{Red} values in parentheses denote the gain of 
w/ PAM over w/o PAM. \textbf{Bold} denotes the better result 
within each subset.}
\label{tab:neg_ratio}
\setlength{\tabcolsep}{7pt}
\begin{tabular}{l|l|ccc|ccc}
\toprule
\multirow{2}{*}{Subset} & \multirow{2}{*}{Method} 
& \multicolumn{3}{c|}{With Constraint} 
& \multicolumn{3}{c}{No Constraint} \\
& & R@10 & R@20 & R@50 & R@10 & R@20 & R@50 \\
\midrule
\multirow{2}{*}{High NI}
& w/o PAM        
& 6.04  & 7.90  & 8.88  
& 5.86  & 8.31  & 11.55 \\
& \cellcolor{gray!15}w/ PAM         
& \cellcolor{gray!15}\gain{\textbf{10.32}}{+4.28}
& \cellcolor{gray!15}\gain{\textbf{12.14}}{+4.24}
& \cellcolor{gray!15}\gain{\textbf{13.97}}{+5.09}
& \cellcolor{gray!15}\gain{\textbf{10.56}}{+4.70}
& \cellcolor{gray!15}\gain{\textbf{13.21}}{+4.90}
& \cellcolor{gray!15}\gain{\textbf{15.79}}{+4.24} \\
\midrule
\multirow{2}{*}{Low NI}
& w/o PAM        
& 33.50 & 39.12 & 40.42 
& 35.00 & 45.17 & 54.30 \\
& \cellcolor{gray!15}w/ PAM         
& \cellcolor{gray!15}\gain{\textbf{33.88}}{+0.38}
& \cellcolor{gray!15}\gain{\textbf{39.89}}{+0.77}
& \cellcolor{gray!15}\gain{\textbf{41.30}}{+0.88}
& \cellcolor{gray!15}\gain{\textbf{35.27}}{+0.27}
& \cellcolor{gray!15}\gain{\textbf{46.11}}{+0.94}
& \cellcolor{gray!15}\gain{\textbf{55.84}}{+1.54} \\
\bottomrule
\end{tabular}
\end{table}
% Neg pair가 reasoning에 악영향을 준다는 정량 결과

\subsection{PAM Margin Sensitivity}
Table~\ref{tab:pam_margin} reports performance under different 
margin settings for $\mathcal{L}_{\text{PAM}}$. Hard margin with 
$m\!=\!1.0$ achieves the best performance, while results remain 
stable across $m \in \{0.5, 1.0, 2.0\}$, indicating robustness 
to the specific margin value. We also evaluate a soft margin 
variant, which replaces the hinge with a log-sum-exp formulation:
\begin{equation}
\mathcal{L}_{\text{PAM}}^{\text{soft}} = \sum_{(a,b^+,b^-)} 
\log\!\left(1 + \exp\!\left(\mathbf{G}_L^{(a,b^-)} - 
\mathbf{G}_L^{(a,b^+)}\right)\right),
\end{equation}
which penalizes all violations proportionally to their magnitude 
rather than enforcing a fixed boundary. Soft margin underperforms 
hard margin variants across most metrics, suggesting that enforcing 
a fixed separation boundary is more effective than a 
distance-proportional penalty for interactive vs. non-interactive 
pair discrimination in our setting.

\begin{table}[h]
\centering
\caption{Sensitivity to PAM margin setting. The shaded row indicates the default setting. \textbf{Bold} and \underline{underline} denote the best and second-best results in each column, respectively.}
\vspace{0.3cm}
\label{tab:pam_margin}
\setlength{\tabcolsep}{8pt}
\begin{tabular}{l|ccc|ccc}
\toprule
Margin & \multicolumn{3}{c|}{With Constraint} 
       & \multicolumn{3}{c}{No Constraint} \\
       & R@10 & R@20 & R@50 & R@10 & R@20 & R@50 \\
\midrule
Soft margin      & 21.79 & 25.63 & 27.33 
                 & \underline{23.69} & 29.94 & 36.31 \\
$m = 0.5$        & 21.92 & 25.77 & \underline{27.40}
                 & 23.67 & 30.15 & 36.59 \\
\rowcolor{gray!15}
$m = 1.0$ (default) & \textbf{22.24} & \textbf{26.48} & \textbf{28.00} 
                    & 23.20 & \underline{30.24} & \textbf{37.47} \\
$m = 2.0$        & \underline{22.01} & \underline{25.84} & 27.36
                 & \textbf{23.73} & \textbf{30.39} & \underline{36.95} \\
\bottomrule
\end{tabular}
\end{table}
% margin m sensitivity, soft margin loss 적용

% ============================================================
%\section{Experiments on VidHOI Dataset}
%\label{sec:vidhoi}
% ============================================================
% VidHOI 데이터셋 설명, 실험 세팅, 결과

% ============================================================
\section{Additional Qualitative Results}
\label{sec:qualitative}
\begin{figure}[h]
    \centering
    \begin{subfigure}{0.76\linewidth}
        \includegraphics[width=\linewidth]{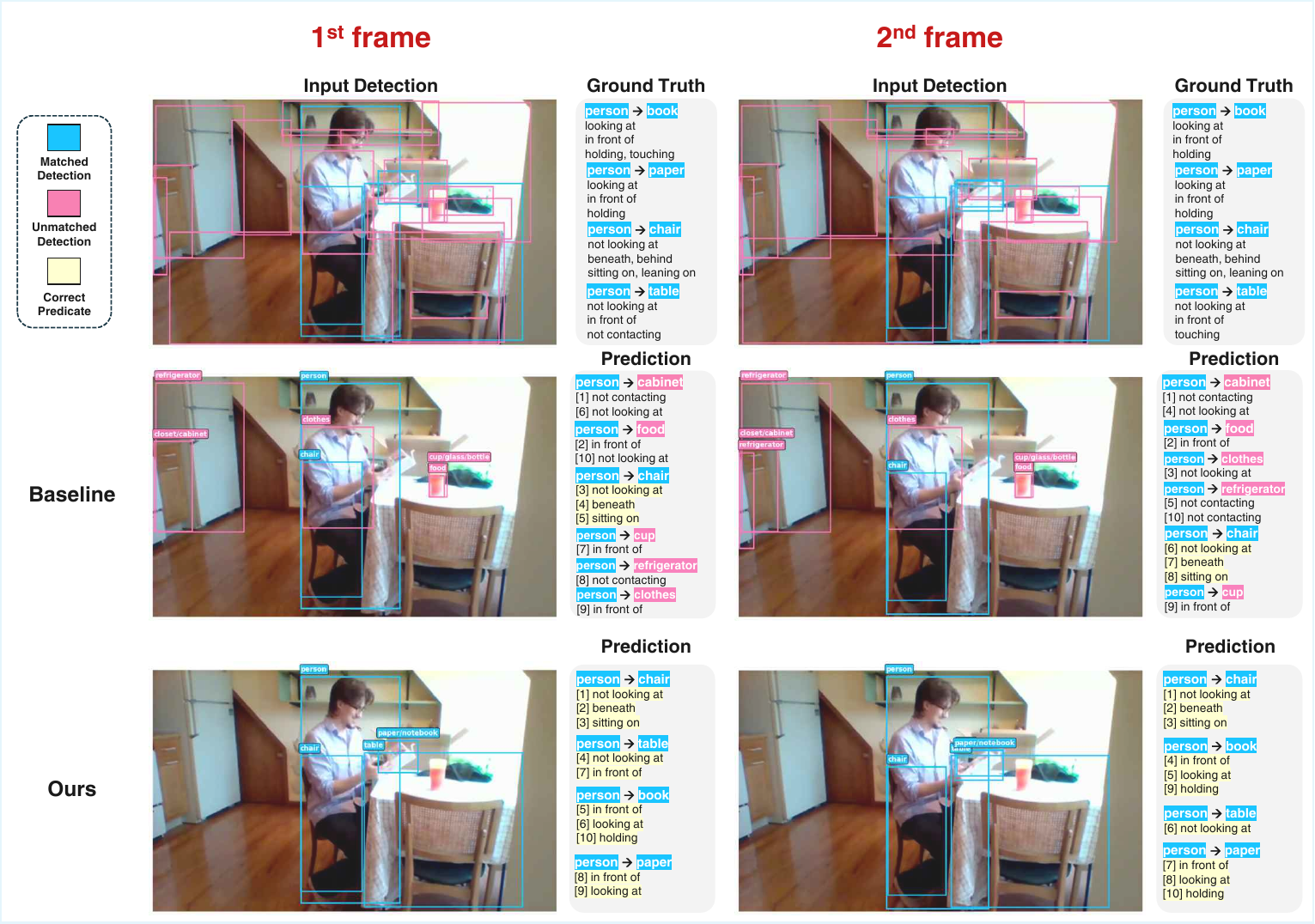}
        \caption{}
        \label{fig:qual1}
    \end{subfigure}
    \vspace{-0.3cm}
    \begin{subfigure}{0.76\linewidth}
        \includegraphics[width=\linewidth]{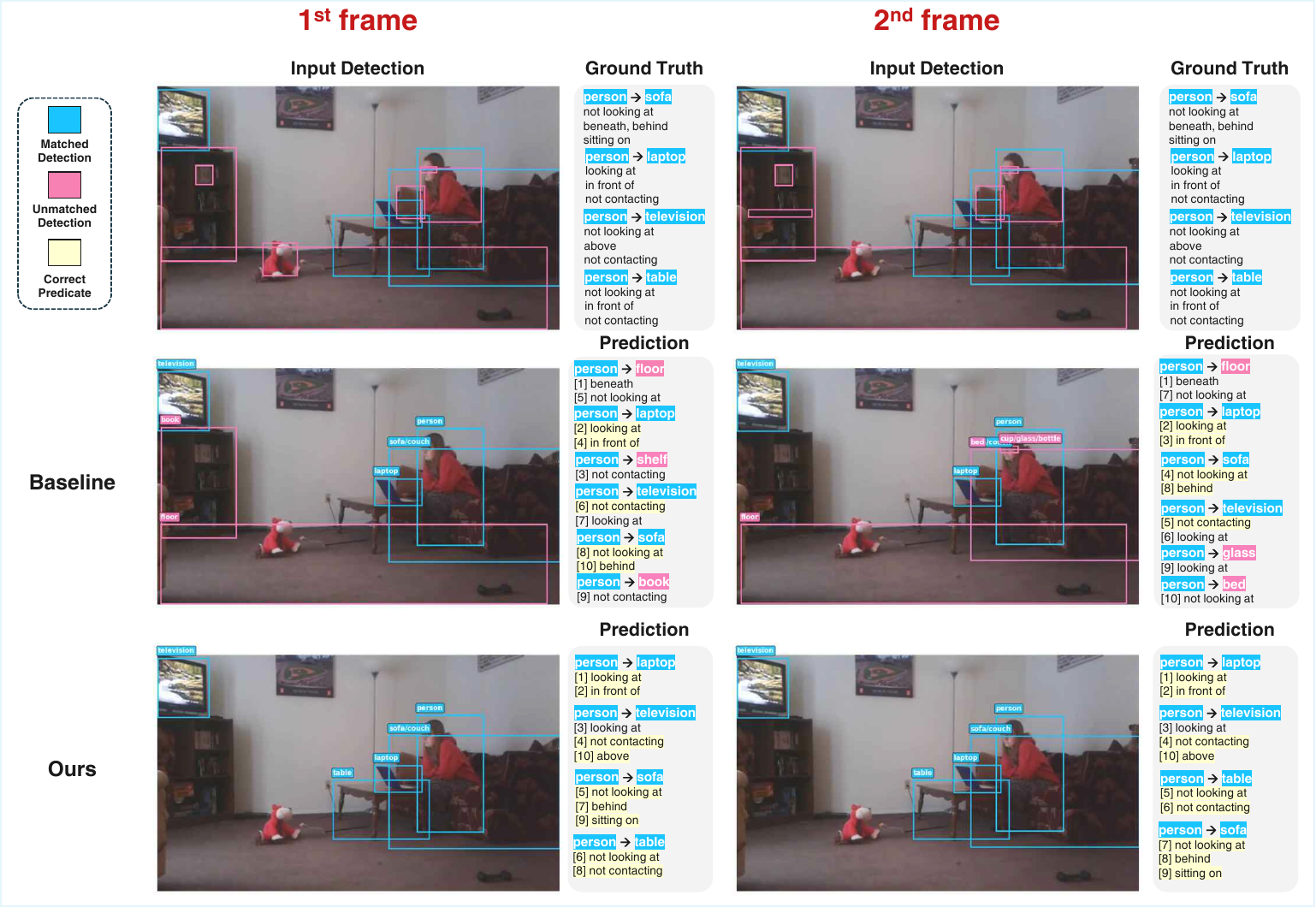}
        \caption{}
        \label{fig:qual2}
    \end{subfigure}
    \caption{Additional qualitative comparisons between the baseline and ours.}
    \label{fig:qualitative}
\end{figure}

We provide additional qualitative comparisons between the 
baseline (PLA with STTran) and our method across two 
video clips from the AG dataset. As shown in 
Figs.~\ref{fig:qual1} and~\ref{fig:qual2}, our method consistently 
produces more accurate and complete scene graph predictions.

\vspace{-0.5cm}

%\bibliographystyle{splncs04}
%\bibliography{main}

%\end{document}